\documentclass[preprint,12pt]{elsarticle}

\usepackage{amssymb}
\usepackage{amsmath}
\usepackage{longtable}
\usepackage[acronym]{glossaries}
\usepackage[nolist,nohyperlinks]{acronym}
\usepackage{caption}
\usepackage{subcaption}
\usepackage{url}
\usepackage{booktabs}
\usepackage{multirow}
\usepackage{graphicx}
\usepackage{pdfpages}
\usepackage{hyperref}

\journal{TBD}

\begin{document}

\begin{frontmatter}


\title{Multi-task learning for the automatic grading of enlarged perivascular space burden using MRI}

\author[inst1,inst2]{Jesse Phitidis}
\author[inst1]{William N. Whiteley}
\author[inst1,inst6]{Joanna M. Wardlaw}
\author[inst7]{Miguel O. Bernabeu}
\author[inst8,inst1]{Yajun Cheng}
\author[inst9,inst1]{Xiaodi Liu}
\author[inst10,inst1]{Junfang Zhang}
\author[inst1,inst6]{Una Clancy}
\author[inst11]{Stephen Makin}
\author[inst1,inst6]{Roberto Duarte Coello}
\author[inst1,inst6]{Susana Mu\~{n}oz Maniega}
\author[inst1]{Mark E. Bastin}
\author[inst12]{Simon R. Cox}
\author[inst1,inst6]{Maria del C. Vald\'{e}s Hern\'{a}ndez\corref{cor1}}
\ead{M.Valdes-Hernan@ed.ac.uk}
\cortext[cor1]{Corresponding author}

\affiliation[inst1]{organization={Centre for Clinical Brain Sciences, University of Edinburgh},
            city={Edinburgh},
            country={UK}}

\affiliation[inst2]{organization={Canon Medical Research Europe},
            city={Edinburgh},
            country={UK}}


\affiliation[inst6]{organization={UK Dementia Research Institute, Centre at The University of Edinburgh},
            city={Edinburgh},
            country={UK}}

\affiliation[inst7]{organization={Usher Institute, University of Edinburgh},
            city={Edinburgh},
            country={UK}}

\affiliation[inst8]{organization={Department of Radiology, Chongqing General Hospital, Chongqing University},
            city={Chongqing},
            country={China}}

\affiliation[inst9]{organization={Department of Psychiatry and Behavioral Sciences, University of California},
            city={San Francisco},
            country={USA}}

\affiliation[inst10]{organization={Department of Neurology, Shanghai Sixth People’s Hospital Affiliated to Shanghai Jiao Tong University School of Medicine},
            city={Shanghai}, 
            country={China}}

\affiliation[inst11]{organization={Centre for Rural Health, University of Aberdeen},
            city={Inverness},
            country={UK}}

\affiliation[inst12]{organization={Department of Psychology, University of Edinburgh},
            city={Edinburgh},
            country={UK}}

\begin{abstract}
Enlarged perivascular spaces (PVS) visible in brain magnetic resonance imaging (MRI) are increasingly thought to be linked to poor brain health. PVS are elongated structures of less than 3 mm in diameter and can be numerous. To reflect the incidence of PVS, radiologists visually score their burden following a clinical grading scale---a task that would benefit from automation to accelerate analyses and overcome the influence of inter-observer differences.
We developed and evaluated methods for training machine learning models to score PVS incidence in the basal ganglia (BG) and centrum semiovale (CSO) leveraging the Potters/Wardlaw scale. The novelty in our work lies in the use of imperfect, semi-automatically generated ``silver-standard'' PVS segmentation masks during training, \textit{in addition} to PVS radiological scores. We comparatively evaluated a conditional convolutional neural network (CNN) which accepts PVS masks as an extra input channel, a multi-task CNN which performs both PVS segmentation and scoring, and a logistic regression model which utilises features derived from PVS masks to predict PVS scores.
Multi-task learning was the most effective method, achieving a mean average precision of 64.08\% compared to 60.22\% for the conditional CNN, 52.11\% for a baseline CNN trained only to predict PVS scores, and 49.32\% for the logistic regression model. The multi-task model showed an ability to localise individual PVS not shown by the other CNNs, and behaved in a probabilistically sensible way, predicting with lower confidence on inherently harder classes. Age, sex, hypertension status, white matter hyperintensity volume, and ischaemic stroke lesion status were shown to be associated with the multi-task model's PVS score predictions and the ground truth in a similar way.
\end{abstract}




\end{frontmatter}


\section{Introduction}


Perivascular spaces (PVS) are fluid-filled cavities surrounding small blood vessels in the brain, which, when they become enlarged, appear visible on magnetic resonance imaging (MRI) scans. They are known to be related to age, vascular risk factors, blood-brain-barrier leakage, cognitive decline, and small vessel disease (SVD) \cite{wardlaw2020perivascular}.

\begin{figure}[htbp]
    \centering
    \includegraphics[width=1.0\linewidth]{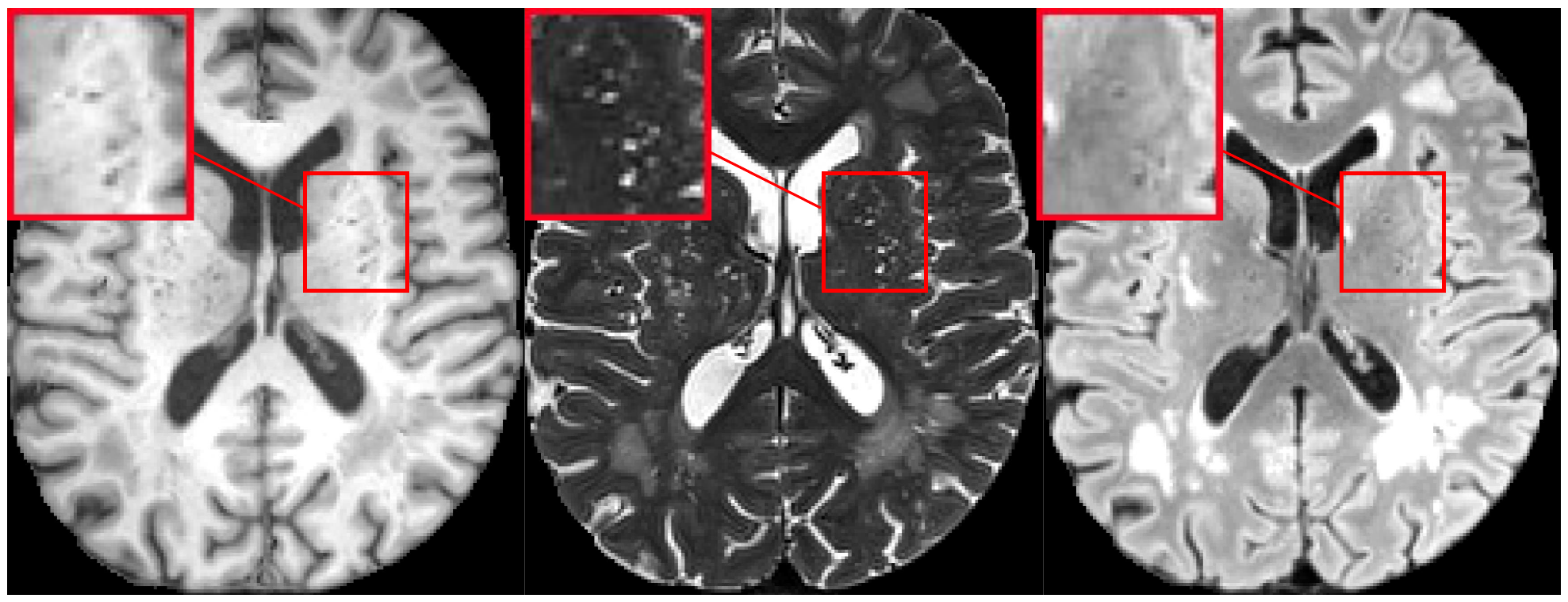}
    \caption{\textbf{Example showing PVS}. T1-weighted (T1w), T2-weighted (T2w), and fluid-attenuated inversion recovery (FLAIR) (left-right) magnetic resonance imaging (MRI) scans. Visible perivascular spaces (PVS) shown in red box, most noticeable as white speckles on T2w scans.}
    \label{ch7_fig:PVS_intro}
\end{figure}

To quantify PVS burden, a number of visual grading scales have been developed. One such scale is the Potters/Wardlaw scale \citep{potter2015cerebral}, which relies on counting visible PVS within the basal ganglia (BG), centrum semiovale (CSO), and midbrain, resulting in one score per region of interest (ROI). The PVS count for one of these regions is defined as the maximum single-hemisphere count over all axial MRI slices where the region is present. For the BG and CSO, the score is then defined as: 0 (0 PVS), 1 (1-10 PVS), 2 (11-20 PVS), 3 (21-40 PVS), or 4 ($\geq$41 PVS). The midbrain follows a binary scale: 0 (0 PVS), and 1 ($\geq$1 PVS). However, in practice, the perceived burden of PVS-like or PVS-associated abnormalities will prevail in assigning the score \cite{gonzalez2017reliability}. 

\vspace{8pt}
While most previous work in automated PVS analysis has focused on segmentation \citep{waymont2024systematic}, there does exist a body of research on automated PVS scoring. 
Regardless of the flooring and ceiling effects that any scoring system may have, disease scoring is currently the most practical way to characterise disease stages in a harmonised way. 
It is also robust against image acquisition differences, especially in the presence of highly anisotropic voxels, prevalent in clinical settings, which are unsuitable for deriving accurate PVS metrics from segmentation methods. 
However, previous automated approaches have generally focused on dichotomised scales or isolated regions. 
For example, \cite{gonzalez2017reliability} developed a support vector machine classifier based on bag of visual words (BoW) features from 2D slices of T2w MRI (masked to the BG), but their model only predicted a dichotomised version of the Potters/Wardlaw PVS score, and only in the BG. 
Similarly, while \cite{williamson2022automated} used a 3D CNN, they also relied on a dichotomised scale for the BG: 0 ($<$10 PVS) or 1 ($\geq$10 PVS). 
Other methods have shifted away from binned ordinal classes entirely; \cite{dubost2019enlarged} trained four independent 3D CNNs to predict continuous PVS counts from T2w MRI (masked to the region of interest) for the BG, CSO, midbrain, and hippocampi, using the scale proposed by \cite{adams2013rating}. 
Even when the standard non-dichotomised Potters/Wardlaw scale has been applied, such as the 2D CNN trained by \cite{yang2021direct} on enhanced and cropped T2w MRIs, predictions have been limited solely to the BG. 
Consequently, a number of gaps remain in the literature. Firstly, no one has yet attempted to automate PVS scoring in both the BG and CSO using a non-dichotomised Potters/Wardlaw scale. Secondly, while the use of PVS score ground truth has previously been used in the development of PVS segmentation methods \citep{ballerini2018perivascular}, PVS segmentations are yet to be utilised in the development of automated PVS scoring models.

\vspace{8pt}
In this work, we develop machine learning models for predicting the BG and CSO PVS scores on a modified (but not dichotomised) Potters/Wardlaw scale. Differently from previous work, we present methods that are designed to utilise PVS segmentation masks (\textit{in addition} to the ground truth PVS scores) during training and/or inference. The masks that we use in training are derived via a semi-automated pipeline and are therefore referred to as ``silver-standard''. The ability to use these masks provides two main benefits. Firstly, the masks provide a lower-level signal, reducing the likelihood of the model learning to predict PVS scores via PVS-correlated imaging features, instead of actual PVS. Secondly, the PVS mask-processing modules can be trained on additional images that lack ground-truth PVS scores, provided that silver-standard PVS masks can be generated for them. This allows these modules (and, indirectly, the overall scoring pipeline) to benefit from a larger and more anatomically and pathologically diverse training set. We emphasise that the silver-standard segmentation masks are imperfect, but cheap to generate compared to either manual segmentations, and even visual PVS scores if datasets are large. Hence, the effective utilisation of these masks represents a real opportunity for scaling up training datasets. Moreover, our approach automatically benefits from any advances in automated PVS segmentation. This work makes the following contributions:

\begin{enumerate}
    \item We develop models for automated BG and CSO PVS scoring on a modified Potters/Wardlaw scale.
    \item We develop and compare three methods for utilising silver-standard PVS segmentation masks \textit{in addition} to ground truth PVS scores in the development of PVS scoring models.
    \item We show that multi-task learning is an effective method of incorporating segmentation supervision into the development of PVS scoring models, yielding a model which is better able to localise individual PVS.
\end{enumerate}

\section{Data}

\subsection{Datasets and ground truth}

We used data from five different datasets (described below, with scanner details in Table \ref{ch7_tab:scanners}), all with T1-weighted (T1w), T2-weighted (T2w), and fluid-attenuated inversion recovery (FLAIR) MRI sequences available. Subjects from 4/5 datasets (i.e., the Mild Stroke Study datasets (MSS1, MSS2, and MSS3) \cite{wardlaw2009lacunar,wardlaw2017blood,clancy2021rationale}, and the Lothian Birth Cohort 1936 dataset (LBC1936) \cite{wardlaw2011brain}) were given a BG and CSO PVS score following the Potters/Wardlaw scale by one of a number of trained observers. Additionally, PVS segmentation masks are available for some subjects. For the MSS1, MSS2, MSS3, and LBC1936 datasets, these PVS masks were generated using a semi-automatic method which utilises vesselness filters and careful threshold selection \cite{valdes2023step,hernandez2024influence} followed by manual edit by a trained analyst. We refer to these masks as ``silver-standard'' because they are not fully manually derived, nor have they all been meticulously scrutinised by an expert. The PVS segmentation masks in the Vascular Lesion Detection Challenge dataset (VALDO) \citep{sudre_where_2024} are manually generated, however there are only six. In the remainder of this paper we refer to all the PVS masks as ``silver-standard'', since the non-VALDO masks constitute the overwhelming majority.

\subsubsection{MSS1}
A cohort of patients presenting with their first clinically evident lacunar or mild cortical stroke to a hospital in the Edinburgh and Lothians regions in Scotland. Recruited prospectively and consecutively at a teaching hospital from 2005-2007. All MRI scans were acquired using a 1.5 Tesla Signa LX General Electric MRI scanner.

\subsubsection{MSS2}
A cohort of patients presenting with clinically evident lacunar or mild cortical stroke, prospectively recruited from 2010-2013. All MRI scans were acquired using a 1.5 Tesla Signa HDxt General Electric MRI scanner.

\subsubsection{MSS3}
A prospective observational cohort study of patients presenting with clinically evident stroke syndromes, beginning in 2018. All MRI scans were acquired using a 3 Tesla Siemens Prisma MRI scanner.

\subsubsection{LBC1936}
A longitudinal observational cohort study of community-dwelling elderly subjects born in the Lothian area in 1936 and living in Edinburgh. All MRI scans were acquired from using a 1.5 Tesla Signa Horizon HDx General Electric MRI scanner from 2007-2010 (``Wave 2'' \citep{taylor2018cohort}).

\subsubsection{VALDO}
Datasets prepared for the ``Where is VALDO'' MICCAI challenge. Data was selected from the SABRE \cite{tillin2012southall,jones2020cohort} sub-dataset of Task 1 (with scans acquired from 2014-2018) and included only the scans with manual PVS segmentation available for the entire image volume.

\begin{table}[htbp]
\centering
\caption{\textbf{MRI scan details}. Magnetic resonance imaging (MRI) scan details (in RAS$+$ coordinate system).}
\label{ch7_tab:scanners}
\begin{tabular}{lllll}
\hline
Dataset                  & Sequence & Type & Field Strength (T) & Resolution (mm)    \\ \hline
\multirow{3}{*}{MSS1}    & T1w      & 2D   & 1.5                & 0.94$\times$0.94$\times$6.50 \\
                         & T2w      & 2D   & 1.5                & 0.94$\times$0.94$\times$6.50 \\
                         & FLAIR    & 2D   & 1.5                & 0.94$\times$0.94$\times$6.50 \\ \hline
\multirow{3}{*}{MSS2}    & T1w      & 3D   & 1.5                & 0.90$\times$1.29$\times$1.29 \\
                         & T2w      & 2D   & 1.5                & 0.47$\times$0.47$\times$6.00 \\
                         & FLAIR    & 2D   & 1.5                & 0.47$\times$0.47$\times$6.00 \\ \hline
\multirow{3}{*}{MSS3}    & T1w      & 3D   & 3.0                & 1.00$\times$1.00$\times$1.00 \\
                         & T2w      & 3D   & 3.0                & 0.90$\times$0.90$\times$0.90 \\
                         & FLAIR    & 3D   & 3.0                & 1.00$\times$1.00$\times$1.00 \\ \hline
\multirow{3}{*}{LBC1936} & T1w      & 3D   & 1.5                & 1.00$\times$1.30$\times$1.00 \\
                         & T2w      & 2D   & 1.5                & 1.00$\times$1.00$\times$2.00 \\
                         & FLAIR    & 2D   & 1.5                & 1.00$\times$1.00$\times$4.00 \\ \hline
\multirow{3}{*}{VALDO}   & T1w      & 3D   & 3.0                & 1.09$\times$1.09$\times$1.00 \\
                         & T2w      & 3D   & 3.0                & 1.09$\times$1.09$\times$1.00 \\
                         & FLAIR    & 3D   & 3.0                & 1.09$\times$1.09$\times$1.00 \\ \hline
\end{tabular}
\end{table}

\subsection{Preprocessing}

\subsubsection{General}
Scans were rigidly registered and resampled to 1 mm isotropic resolution. We used SynthStrip \citep{hoopes2022synthstrip} for brain extraction (using the T2w MRI scan for each subject). Scans were then cropped to the brain bounding box, before zero-padding to ensure the image shape was a multiple of 32. Finally, intensity normalisation was applied within the brain mask, saturating at the 2nd and 98th percentile, and mapping all values into the range [0,1].

\subsubsection{Generating region of interest masks}
To generate binary masks for the BG and CSO, we combined anatomical segmentations from SynthSeg \citep{billot2023synthseg} with a manually drawn exclusion template (a binary mask covering midline and ventricular-adjacent areas where artefacts frequently mimic PVS). We non-linearly registered this custom template to each subject’s T1w image using NiftyReg \citep{modat2010fast, modat2014global}.

For the BG mask, we started by defining BG-relevant anatomies using the SynthSeg output. These regions are the thalamus, caudate, putamen, pallidum, and accumbens area. Because these initial labels can be overly conservative for the BG, we expanded the region using a 6 mm binary dilation. To ensure this expansion did not encompass unwanted structures, we removed any voxels overlapping with the registered exclusion template. Finally, we constrained the BG mask to voxels within either the initial BG-relevant anatomies, or cerebral white matter.

\begin{figure}
     \centering
     \begin{subfigure}[b]{0.49\textwidth}
         \centering
         \includegraphics[width=\textwidth]{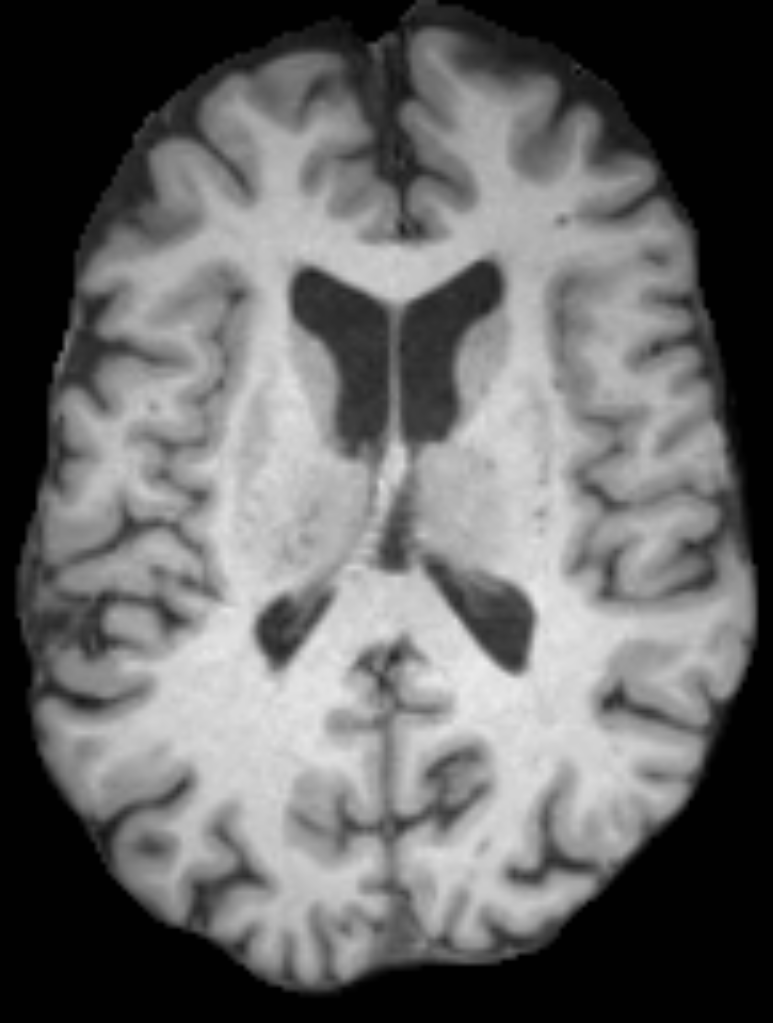}
     \end{subfigure}
     \hfill
     \begin{subfigure}[b]{0.49\textwidth}
         \centering
         \includegraphics[width=\textwidth]{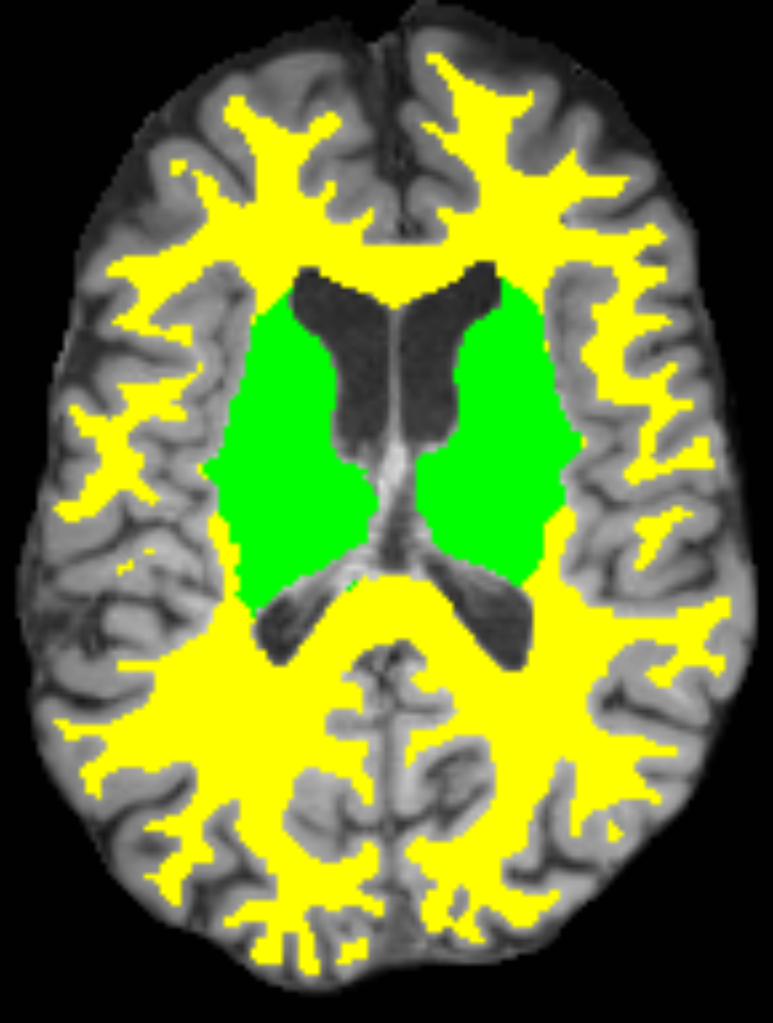}
     \end{subfigure}
     \caption{\textbf{Example of generated region of interest masks}. T1-weighted (T1w) magnetic resonance imaging (MRI) scan showing basal ganglia (BG) (green) and centrum semiovale (CSO) (yellow) masks.}
     \label{ch7_fig:BG_CSO_masks}
\end{figure}

For the CSO mask, we initialised a mask using the cerebral white matter labels produced by SynthSeg. We then subtracted any voxels that intersected with either the exclusion template or the finalised BG mask. Figure \ref{ch7_fig:BG_CSO_masks} shows an example of the resulting masks.

\subsection{Agreement between PVS masks and PVS scores} \label{ch7_sec:agreement}

Given the unambiguous nature of the Potters/Wardlaw scale as it is described in the user guide\footnote{\url{https://edwebcontent.ed.ac.uk/sites/default/files/imports/fileManager/epvs-rating-scale-user-guide.pdf}}, in theory, solving PVS segmentation would also solve PVS scoring. We derived maximum PVS counts from the silver-standard PVS segmentation maps, algorithmically, by looping over every slice and each hemisphere, performing 2D connected components analysis, and counting the number of components within the region of interest (BG or CSO) mask in the slice. The maximum number recorded in each ROI is then, in theory, the number that a radiologist would use to assign a score on the Potters/Wardlaw scale. 

\begin{figure}[htbp]
     \centering
     \begin{subfigure}[b]{0.49\textwidth}
         \centering
         \includegraphics[width=\textwidth]{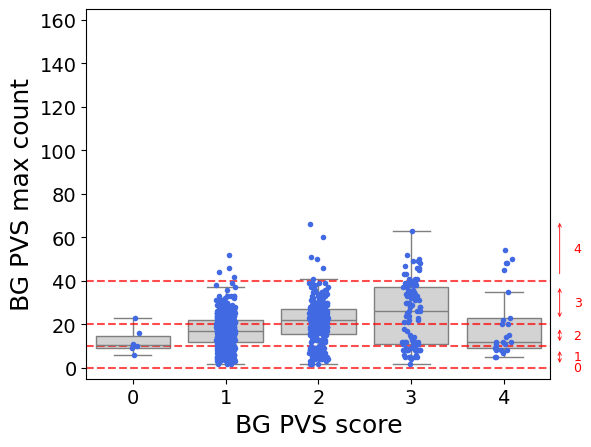}
     \end{subfigure}
     \hfill
     \begin{subfigure}[b]{0.49\textwidth}
         \centering
         \includegraphics[width=\textwidth]{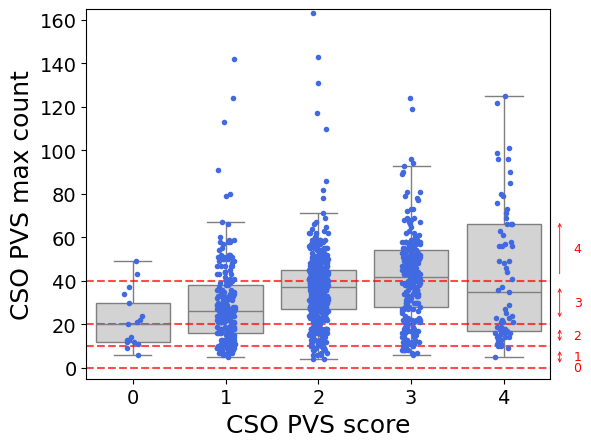}
     \end{subfigure}
     \caption{\textbf{Figure showing disagreement between PVS scores and masks}. Distribution of maximum perivascular space (PVS) counts (as derived from silver-standard segmentation masks using the Potters/Wardlaw counting criteria) for subjects with different ground truth scores (as rated by a trained observer). Dashed red lines show the count thresholds at which the score on the Potters/Wardlaw scale would be expected to change, with score 0 being on the first line, and score 4 being above the last line. The results show very high disagreement between the counts and the scores for the basal ganglia (BG) and centrum semiovale (CSO). Box plot whiskers extend to 1.5 times the interquartile range.}
     \label{ch7_fig:BG_and_CSO_box}
\end{figure}

Figure \ref{ch7_fig:BG_and_CSO_box} shows that, for our data, this algorithmic segmentation-based approach to PVS scoring does not work sufficiently well. Notably, the segmentation-derived maximum PVS count is often much higher, especially for CSO, than would be expected given the visual PVS score. There are two reasons as to why these differences may arise. The first reason is related to the segmentation masks themselves; as noted already, these masks are not expertly manually annotated and so, while they have been confirmed to be of a reasonable quality, they are not perfect. The second reason is related to the scores. While the Potters/Wardlaw scale is perfectly clear, it is unrealistic to expect a busy clinical researcher to painstakingly count the number of PVS in each slice and each hemisphere of an MRI scan. It is much more realistic for them to examine the MRI scan, with the Potters/Wardlaw scale in mind, and give their ``impression'' of the score based on the perceived severity. This ``impression'' is extremely valuable, since it may take into account additional factors (e.g.,~other small vessel disease markers, scan quality, etc.). For this reason, learning a model which can approximate this expert ``impression'' is the ultimate goal of this research and we consider the PVS score the ultimate ground truth, while the silver-standard segmentation masks should be thought of as additional and correlated data that we can use to ground our PVS scoring models via segmentation-aware processing.

\subsection{Data splits and modified PVS score}

We split our dataset into training (70\%), validation (10\%), and test (20\%) sets, stratifying by BG and CSO PVS score. The six VALDO cases which did not have scores were added to the training set as well, and used as additional training samples for the segmentation U-Net and the multi-task CNN (described in Section \ref{ch7_sec:methods}). Table \ref{ch7_tab:data_tab} shows the number of subjects from each dataset in each training split.

\begin{table}[htbp]
\centering
\caption{\textbf{Availability of PVS scores and masks in our dataset}. Table showing the number of subjects from each dataset and each data split with available silver-standard perivascular space (PVS) segmentation masks and ground truth PVS scores.}
\label{ch7_tab:data_tab}
\resizebox{\textwidth}{!}{%
\begin{tabular}{l|ccc|ccc|ccc}
\hline
\multirow{2}{*}{Dataset} & \multicolumn{3}{c|}{Training}        & \multicolumn{3}{c|}{Validation}      & \multicolumn{3}{c}{Test}             \\ \cline{2-10} 
 & Masks & \multicolumn{1}{c|}{Scores} & Total & Masks & \multicolumn{1}{c|}{Scores} & Total & Masks & \multicolumn{1}{c|}{Scores} & Total \\ \hline
MSS1                     & 0   & \multicolumn{1}{c|}{67}  & 67  & 0   & \multicolumn{1}{c|}{8}   & 8   & 0   & \multicolumn{1}{c|}{18}  & 18  \\
MSS2                     & 177 & \multicolumn{1}{c|}{177} & 178 & 17  & \multicolumn{1}{c|}{17}  & 17  & 61  & \multicolumn{1}{c|}{62}  & 62  \\
MSS3                     & 160 & \multicolumn{1}{c|}{160} & 160 & 28  & \multicolumn{1}{c|}{28}  & 28  & 40  & \multicolumn{1}{c|}{40}  & 40  \\
LBC1936                  & 346 & \multicolumn{1}{c|}{463} & 463 & 59  & \multicolumn{1}{c|}{71}  & 71  & 96  & \multicolumn{1}{c|}{128} & 128 \\
VALDO                    & 6   & \multicolumn{1}{c|}{0}   & 6   & 0   & \multicolumn{1}{c|}{0}   & 0   & 0   & \multicolumn{1}{c|}{0}   & 0   \\ \hline
Total                    & 689 & \multicolumn{1}{c|}{867} & 874 & 104 & \multicolumn{1}{c|}{124} & 124 & 197 & \multicolumn{1}{c|}{248} & 248 \\ \hline
\end{tabular}%
}
\end{table}

\begin{figure}[htbp]
     \centering
     \begin{subfigure}[b]{0.49\textwidth}
         \centering
         \includegraphics[width=\textwidth]{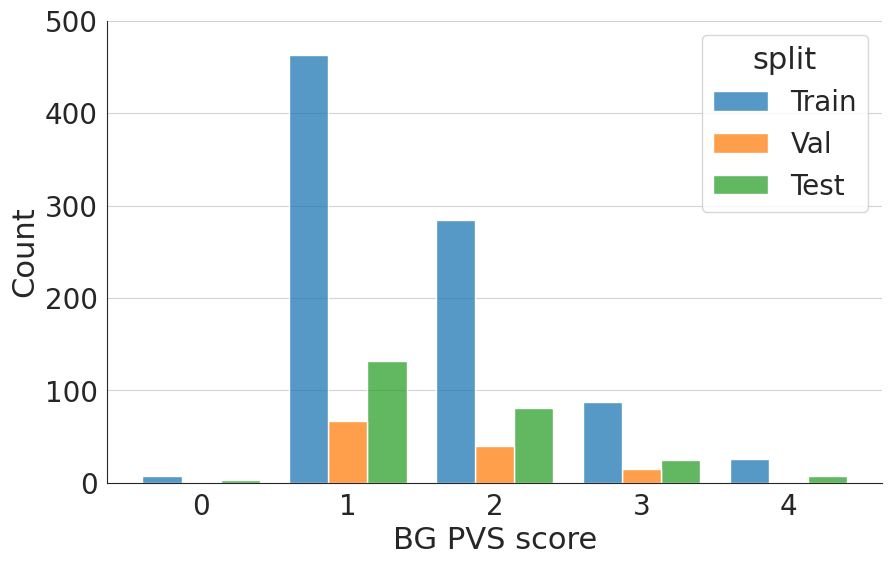}
     \end{subfigure}
     \hfill
     \begin{subfigure}[b]{0.49\textwidth}
         \centering
         \includegraphics[width=\textwidth]{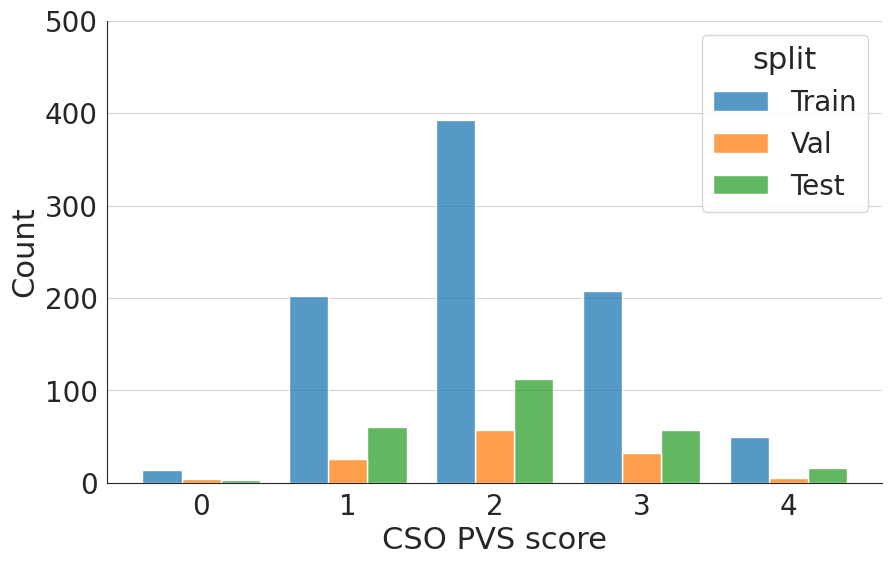}
     \end{subfigure}
     \hfill
     \begin{subfigure}[b]{0.49\textwidth}
         \centering
         \includegraphics[width=\textwidth]{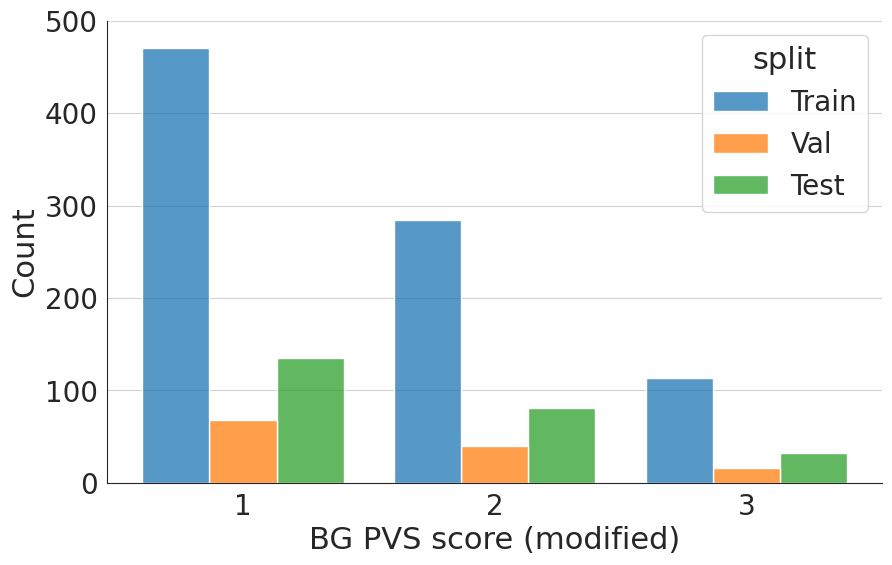}
     \end{subfigure}
     \hfill
     \begin{subfigure}[b]{0.49\textwidth}
         \centering
         \includegraphics[width=\textwidth]{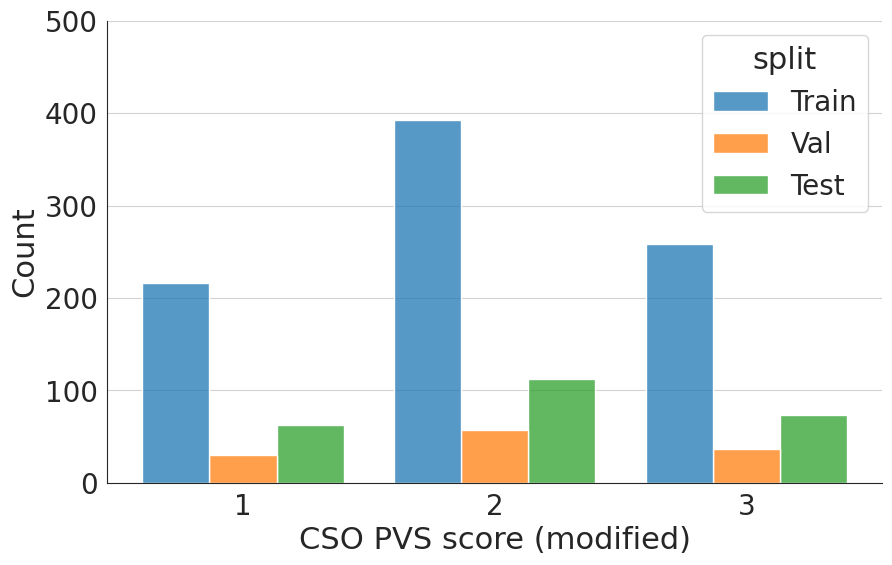}
     \end{subfigure}
     \caption{\textbf{Histograms of PVS scores in our dataset}. \textbf{Top}: the frequency of basal ganglia (BG) and centrum semiovale (CSO) perivascular space (PVS) scores. \textbf{Bottom}: the frequency of our modified BG and CSO PVS scores. The modified scores merge 0 into 1 and 4 into 3.}
     \label{ch7_fig:BG_and_CSO_hist}
\end{figure}

Figure \ref{ch7_fig:BG_and_CSO_hist} shows the frequency of BG and CSO PVS scores in our dataset. Due to the very low number of subjects with scores of 0 or 4, we elected to merge 0 into 1 and 4 into 3. We hope that any modelling success achieved using our modified scale could be replicated using the full Potters/Wardlaw scale, given enough data points. The frequency of our modified scores is also shown in figure \ref{ch7_fig:BG_and_CSO_hist}.

\section{Methods} \label{ch7_sec:methods}


We developed and evaluated automatic methods for predicting the PVS score in the BG and CSO, according to the Potters/Wardlaw scale. In this section, we describe these methods, illustrated in Figure \ref{ch7_fig:one}, at a high level. Much more detailed descriptions are provided in \ref{ch7_app:configs}. Code is available at: \url{https://github.com/Jesse-Phitidis/PVS_SCORING}

\begin{figure}
    \centering
    \includegraphics[width=0.78\linewidth]{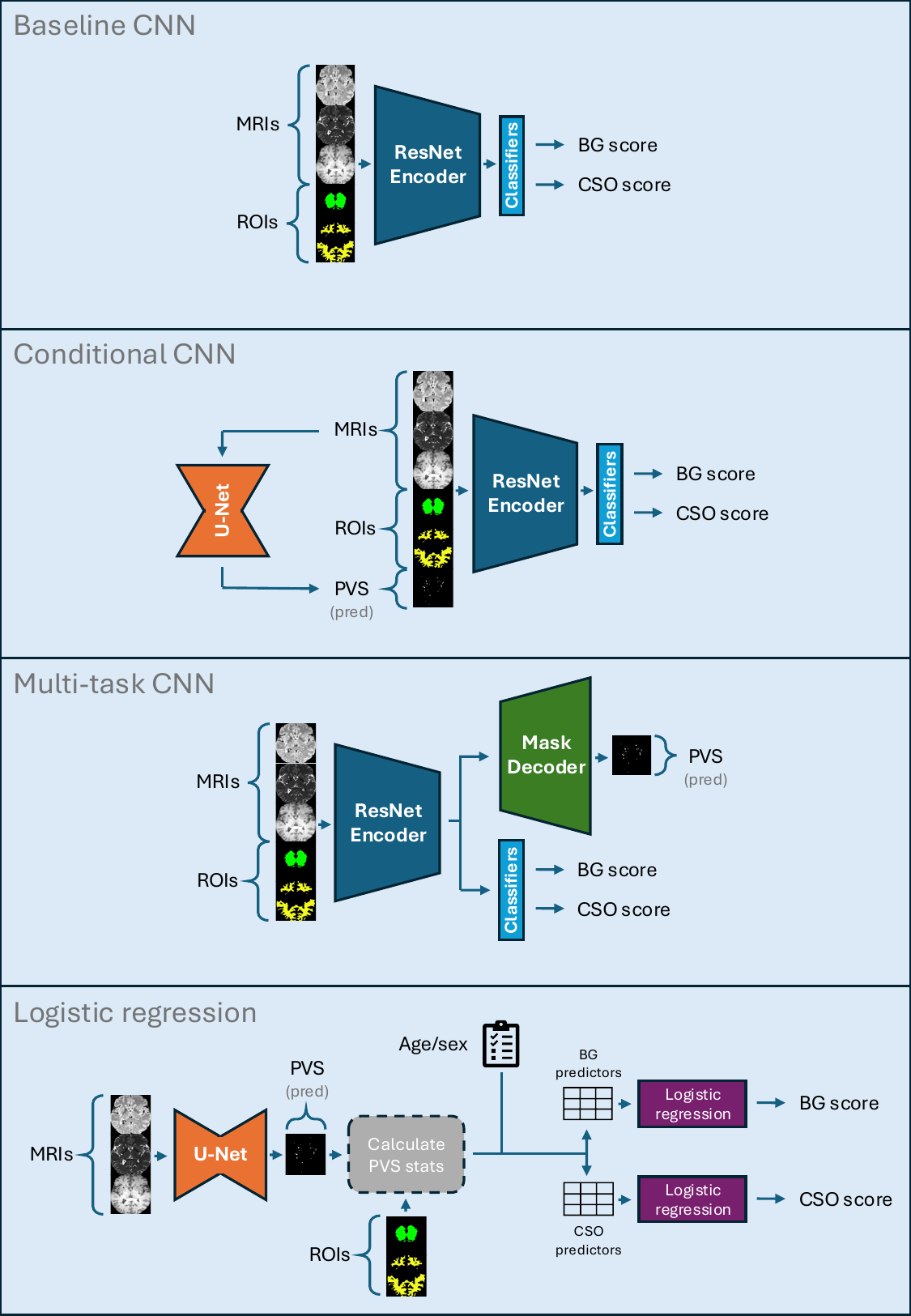}
    \caption{\textbf{Overview of the proposed PVS score prediction methods}. \textbf{Baseline CNN}: trained to classify basal ganglia (BG) and centrum semiovale (CSO) perivascular space (PVS) scores. \textbf{Conditional CNN}: trained to classify BG and CSO PVS scores while utilising PVS segmentation predictions from a U-Net that is pre-trained on the subset of data with silver-standard PVS segmentation masks. \textbf{Multi-task CNN}: trained to classify BG and CSO PVS scores and to predict the silver-standard PVS segmentation masks. \textbf{Logistic regression}: Extracts statistical features from the pre-trained U-Net's PVS segmentation predictions (e.g.,~volumes/count) and combines these with age/sex to form two sets of predictor variables (one for the BG model and one for CSO model), with feature selection on the validation set. CNN: convolutional neural network.}
    \label{ch7_fig:one}
\end{figure}

\subsection{Segmentation U-Net}

We trained a 3D U-Net model \citep{ronneberger2015u} to segment PVS from T1w, T2w, and FLAIR MRI sequences, using the subset of our training data which contains silver-standard PVS segmentation masks. Once trained, this provides us with a fast way to produce PVS probability maps for any new scan, facilitating downstream PVS score prediction. The agreement between the model's predicted PVS masks and the silver-standard PVS masks is evaluated in Section \ref{ch7_sec:results_seg}, and the effect of using the predictions in place of the original masks in downstream PVS score prediction is ablated in \ref{ch7_app:abl}. The model was trained using Dice loss ($\mathcal{L}_{Dice}$) and Cross-Entropy loss ($\mathcal{L}_{CE}$) loss. The overall segmentation loss ($\mathcal{L}_{seg}$) is hence given by:

\begin{equation}
    \mathcal{L}_{seg} = \mathcal{L}_{Dice} + \mathcal{L}_{CE}.
\end{equation}

\subsection{Baseline CNN}

As a baseline, we trained a 3D ResNet \citep{he2016deep} for PVS score prediction. This method does not utilise the segmentation masks available in the training dataset. It takes the multimodal MRI scans (T1w, T2w, FLAIR) as input, in addition to the BG and CSO binary masks. It outputs two three-class probability distributions; one for the predicted BG PVS score and one for the predicted CSO PVS score. The model was trained using Cross-Entropy loss. The classification loss ($\mathcal{L}_{cls}$) is given by:

\begin{equation}
    \mathcal{L}_{cls} = 0.5 \mathcal{L}^{\text{BG}}_{CE} + 0.5 \mathcal{L}^{\text{CSO}}_{CE}
\end{equation}

\subsection{Conditional CNN}

This method follows an identical setup to the baseline CNN, with the exception of the model's input. An additional input channel is added, which processes the soft predictions from the segmentation U-Net. Prior to training the conditional CNN, the segmentation U-Net is run on the subset of the training data which contains PVS score ground truth. For the subjects which did not have silver-standard binary PVS masks originally, we now attain soft pseudo-labels. For the subjects which did originally have silver-standard PVS masks, we also produce soft pseudo-labels (i.e.,~the segmentation U-Net predicts on its own training data).

\subsection{Multi-task CNN}

This method employs the same 3D ResNet encoder as the baseline and conditional CNNs, but with two separate task heads. The first task head is the normal classification layer from the baseline and conditional CNNs, while the second is a U-Net-like mask decoder, which predicts the segmentation mask. The model is trained to perform both tasks simultaneously. The multi-task loss ($\mathcal{L}_{multi}$) is a weighted sum of the segmentation and classification loss components and is given by
\begin{equation}
    \mathcal{L}_{multi} = \alpha \beta \mathcal{L}_{seg} + (1 - \alpha)(2 - \beta) \mathcal{L}_{cls}
\end{equation}

where $\alpha$ is a hyperparameter and, because not all subjects have ground truth for both tasks, $\beta$ is a normalising constant designed to ensure an equal baseline contribution over a full training epoch, and is defined as 

\begin{equation}
    \beta = \frac{2N_{cls}}{N_{cls}+N_{seg}}
\end{equation}

where $N_{cls}$ and $N_{seg}$ are the number of scans with PVS score ground truth and PVS segmentation masks respectively.

\subsection{Logistic regression}
This modelling approach uses two multinomial logistic regression models; one to predict the BG PVS score and one to predict the CSO PVS score. For each model, we performed feature and kernel selection on the validation set, using the mean average precision (mAP) to evaluate performance. Potential features included age, sex, and features derived from the binarised PVS pseudo-labels generated by the segmentation U-Net. Polynomial kernels of degrees 1-3 were evaluated. The selected model for BG PVS prediction utilised a degree 2 polynomial kernel with three features: (1) age; (2) BG PVS volume in the whole scan; and (3) the maximum count of BG PVS in one hemisphere of any slice. The best model for CSO PVS prediction utilised a degree 3 polynomial with three features: (1) sex; (2) the total count of PVS in the whole scan; and (3) the maximum volume of CSO PVS in one hemisphere of any slice.

\subsection{Random seeds and model checkpoints}

We trained the baseline, conditional, and multi-task CNNs three times, using three different random seeds for initialisation, and performed inference by way of ensembling the probabilistic predictions from each model. The segmentation U-Net used a single random seed. Logistic regression model fitting is a convex optimisation problem, hence, no ensemble was used for this method.

For the segmentation U-Net, we selected the model checkpoint achieving the highest Dice similarity coefficient (DSC) \citep{dice1945measures} on the validation set, and for the baseline, conditional, and multi-task CNNs we selected the model checkpoints achieving the highest mAP on the validation set.

\subsection{Evaluation metrics}

To evaluate the segmentation performance of the U-Net and multi-task CNNs, we utilised the Dice similarity coefficient (DSC) metric, which measures the overlap between a ground truth mask $Y$ and predicted mask $\hat{Y}$, given by:

\begin{equation}
    \text{DSC} = \frac{2|Y \cap \hat{Y}|}{|Y| + |\hat{Y}|}.
\end{equation}

To evaluate PVS classification performance, we employed three metrics: mean average precision (mAP), macro-average F1 score (F1), and accuracy (ACC). The average precision (AP) is also known as the area under the precision recall curve (AUPRC) and is a threshold-independent metric. For each class ($c$) of the $C$ total classes, it is calculated as a weighted sum of the precision ($P_t^{(c)}$) over all thresholds ($t$) (out of the $T$ thresholds which result in changes to the prediction), using the increase in recall ($R_t^{(c)}$) between each threshold as the weight:

\begin{equation}
    \text{mAP} = \frac{1}{C}\sum_c^{C} \sum_t^T (R_t^{(c)} - R_{t-1}^{(c)}) P_t^{(c)}.
\end{equation}

The F1 score is mathematically identical to the DSC, but is often described in terms of the number of true positives ($\text{TP}^{(c)}$), false positives ($\text{FP}^{(c)}$), and false negatives ($\text{FN}^{(c)}$) in the context of classification. The macro-average F1 score is:

\begin{equation}
    \text{F1} = \frac{1}{C} \sum_c^C \frac{2\text{TP}^{(c)}}{2\text{TP}^{(c)} + \text{FP}^{(c)} + \text{FN}^{(c)}}.
\end{equation}

Finally, the accuracy (ACC) is simply the fraction of correct predictions.

For the PVS classification metrics, we performed 10,000 bootstrap samplings of the test set to calculate 95\% confidence intervals. We used the distribution of pairwise differences in metric results over these bootstrap iterations to calculate p-values for non-zero differences, and corrected for multiple comparisons using Bonferroni correction ($n=6$).

\section{Results and discussion}

\subsection{Segmentation model validation} \label{ch7_sec:results_seg}

\begin{figure}[htbp]
     \centering
     \begin{subfigure}[b]{0.49\textwidth}
         \centering
         \includegraphics[width=\textwidth]{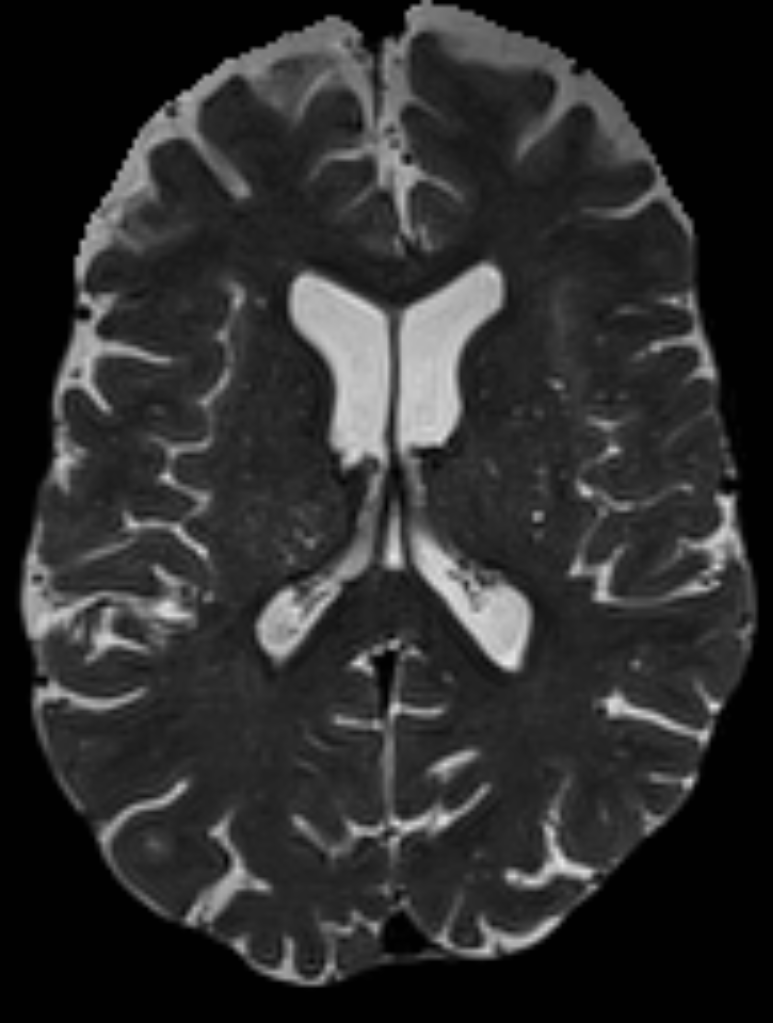}
     \end{subfigure}
     \hfill
     \begin{subfigure}[b]{0.49\textwidth}
         \centering
         \includegraphics[width=\textwidth]{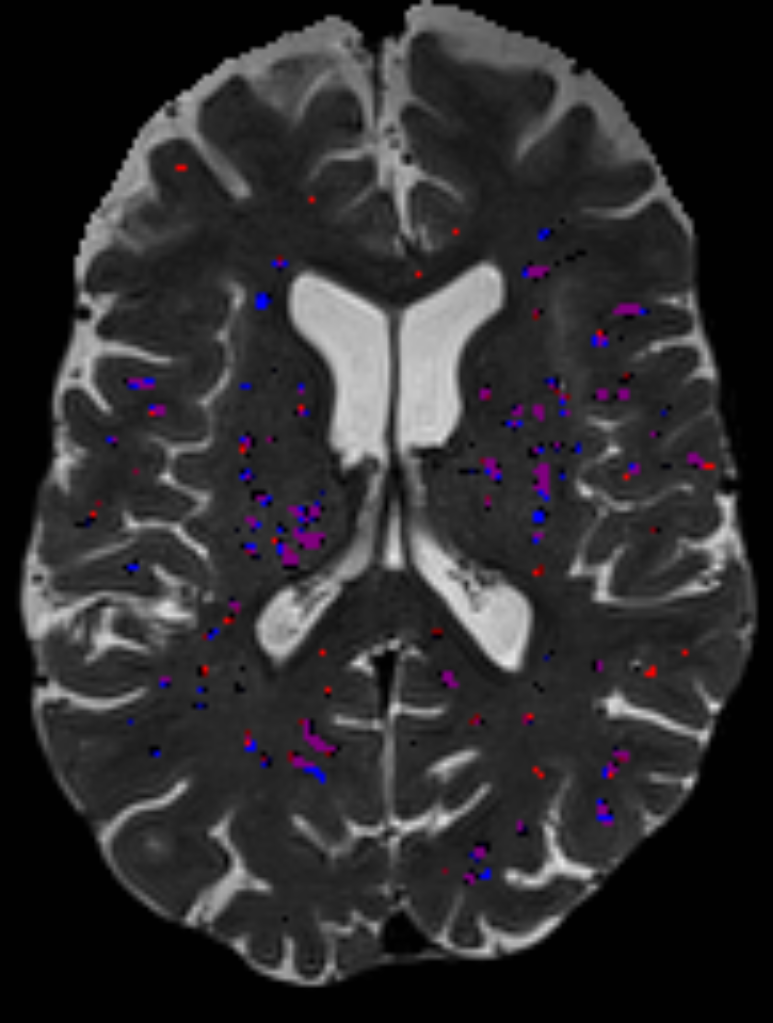}
     \end{subfigure}
     \caption{\textbf{Qualitative evaluation of the U-Net's PVS mask predictions}. T2-weighted (T2w) magnetic resonance imaging (MRI) scan from test set and segmentation overlays. Red shows the silver-standard perivascular space (PVS) mask and blue shows the segmentation U-Net's predicted PVS mask, with purple showing overlap.}
     \label{ch7_fig:pvs_seg_vis}
\end{figure}

We trained our segmentation U-Net to approximate the silver-standard PVS segmentation masks. Since these masks were produced by a slower approach, requiring manual tuning, this was necessary to enable our PVS scoring pipelines to be deployed efficiently on new scans. In this section, we evaluate the agreement between the silver-standard PVS masks, and our trained segmentation U-Net model's predictions. The segmentation U-Net achieved a DSC of 65.06\%, 63.45\%, and 62.84\% on the training, validation, and test sets, respectively, when compared to the silver-standard PVS masks. An example of the PVS segmentations can be seen in Figure \ref{ch7_fig:pvs_seg_vis}. 

\begin{figure}[htbp]
     \centering
     \begin{subfigure}[b]{0.49\textwidth}
         \centering
         \includegraphics[width=\textwidth]{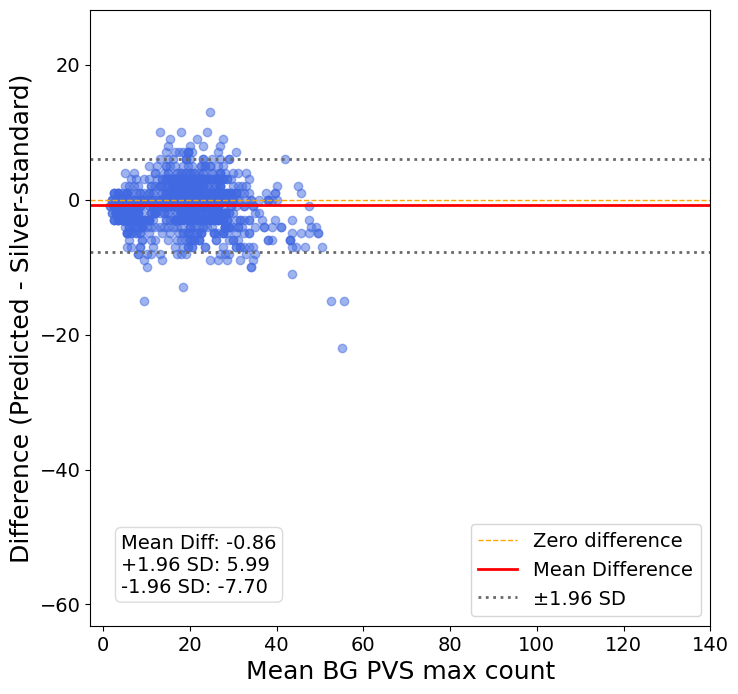}
     \end{subfigure}
     \hfill
     \begin{subfigure}[b]{0.49\textwidth}
         \centering
         \includegraphics[width=\textwidth]{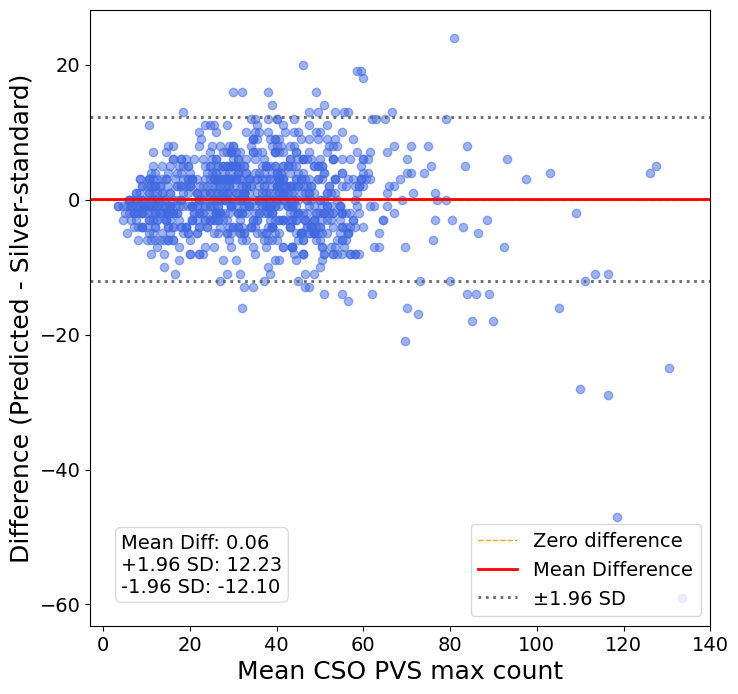}
     \end{subfigure}
     \caption{\textbf{Volumetric analysis of the U-Net's PVS mask predictions}. Bland-Altman plots showing maximum PVS counts within the BG and CSO (according to the Potters/Wardlaw counting criteria) as derived from the silver-standard perivascular space (PVS) masks and the PVS masks predicted by the segmentation U-Net. Data points include all training, validation, and test subjects with silver-standard PVS masks available.}
     \label{ch7_fig:pred_gt_agreement}
\end{figure}

As in Section \ref{ch7_sec:agreement}, we derived the maximum PVS count for each subject according to the silver-standard PVS segmentation masks. We compared this with the count derived from the predicted PVS masks. The results in Figure \ref{ch7_fig:pred_gt_agreement} show that for both BG and CSO counts, the mean difference is less than one, although the confidence intervals are larger for CSO due to the higher PVS counts from both segmentation masks.

While the actual agreement of the two segmentation masks is interesting, we are most concerned with assessing whether or not the segmentation U-Net's masks possess similar utility to the silver-standard masks, when used in downstream PVS score prediction pipelines. This is indeed confirmed to be the case for the conditional CNN and logistic regression methods. These experiments are presented in \ref{ch7_app:abl}.

It is also interesting to evaluate the performance of the multi-task CNN on PVS segmentation. We evaluated the DSC compared with the silver-standard masks for each of the three models (from the three random seeds) and averaged the result. The model achieved 55.81\%, 56.88\%, and 54.27\% on the training, validation, and test sets, respectively. This is considerably lower than the segmentation U-Net, as expected, and we attribute the result to three primary factors: (1) the architecture is not optimised for segmentation (the mask decoder has less parameters than the U-Net's decoder, deep supervision is not used, and group normalisation is used instead of instance normalisation); (2) while we show that the segmentation task is beneficial to the model's ability to detect and score PVS (as indicated by the results below), this is likely not a mutually beneficial relationship; and (3) we selected the checkpoints based on mAP on the PVS scoring task, not DSC on the PVS segmentation task.

\subsection{Comparison of PVS scoring models}

\begin{figure}[htbp]
     \centering
     \begin{subfigure}[b]{\textwidth}
         \centering
         \includegraphics[width=0.25\textwidth]{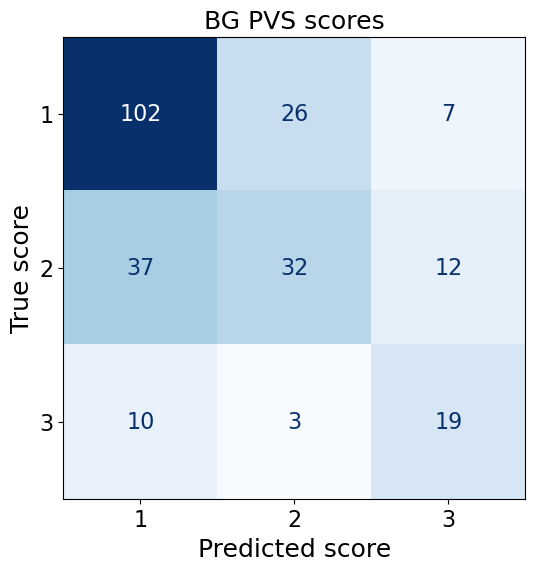}
         \includegraphics[width=0.25\textwidth]{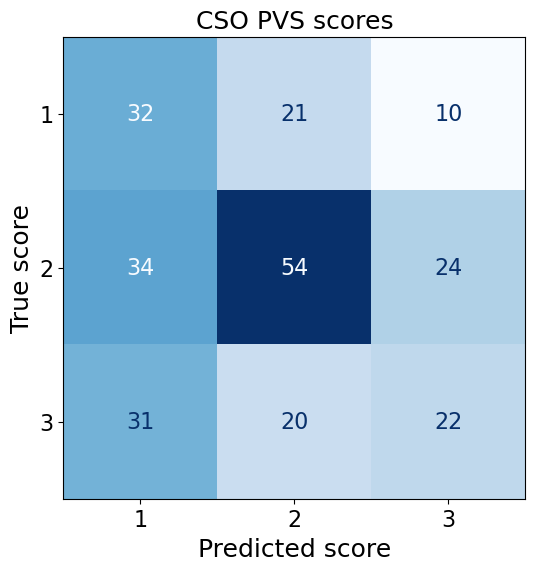}
         \caption{Logistic regression}
     \end{subfigure}
     \hfill
     \begin{subfigure}[b]{\textwidth}
         \centering
         \includegraphics[width=0.25\textwidth]{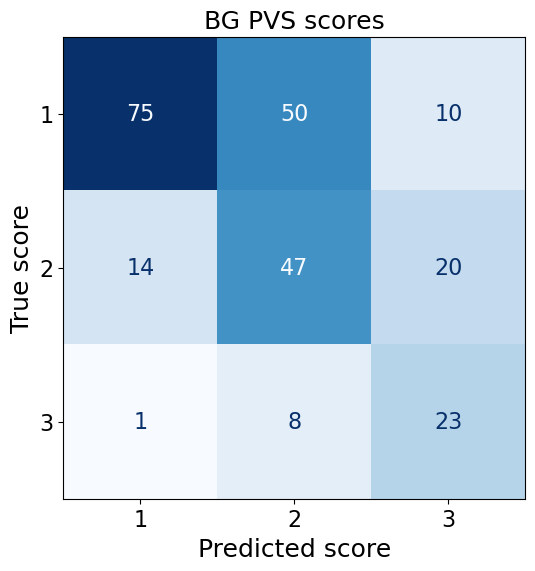}
         \includegraphics[width=0.25\textwidth]{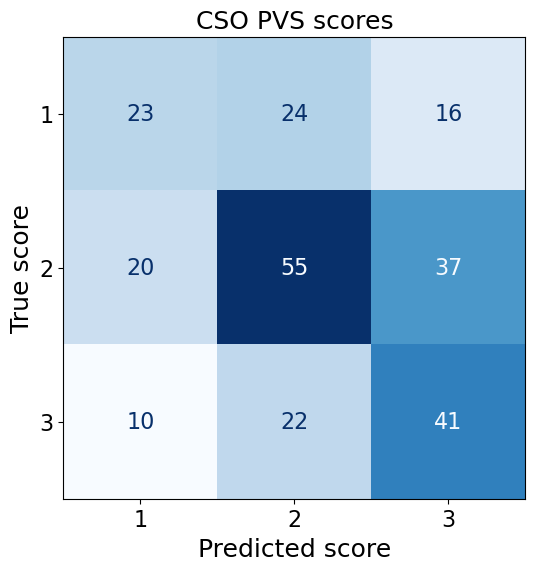}
         \caption{Baseline CNN}
     \end{subfigure}
     \hfill
     \begin{subfigure}[b]{\textwidth}
         \centering
         \includegraphics[width=0.25\textwidth]{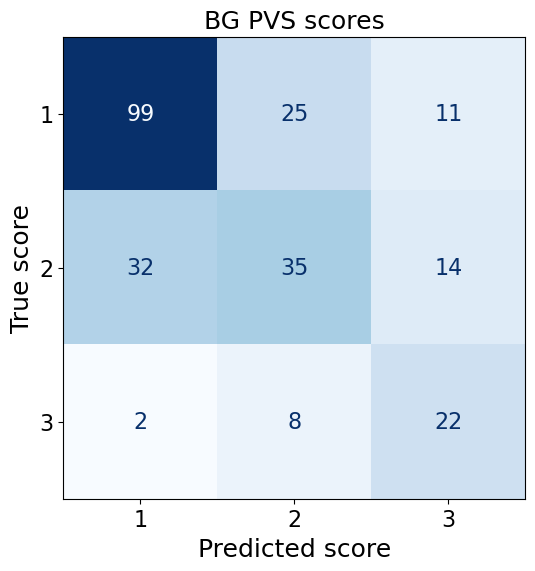}
         \includegraphics[width=0.25\textwidth]{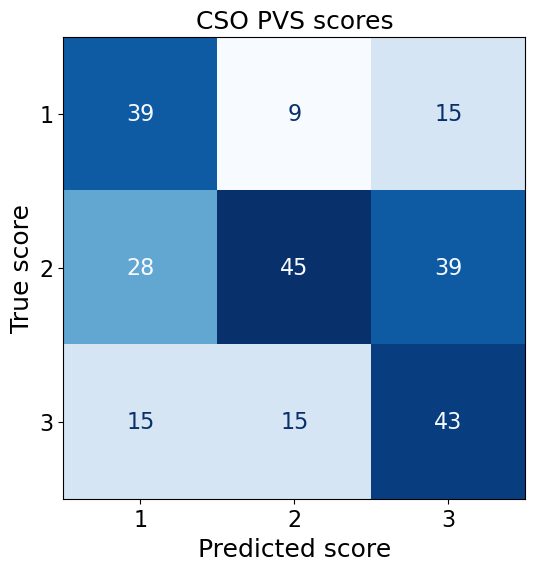}
         \caption{Conditional CNN}
     \end{subfigure}
     \hfill
     \begin{subfigure}[b]{\textwidth}
         \centering
         \includegraphics[width=0.25\textwidth]{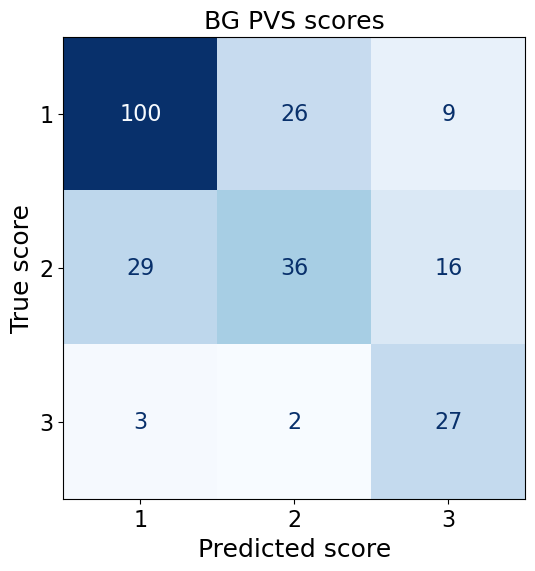}
         \includegraphics[width=0.25\textwidth]{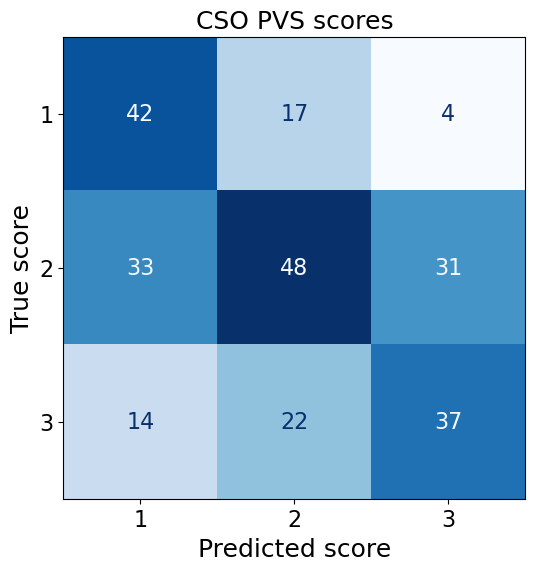}
         \caption{Multi-task CNN}
     \end{subfigure}
     \caption{\textbf{Confusion matrices}. Basal ganglia (BG) and centrum semiovale (CSO) perivascular space (PVS) score predictions, on the test set. CNN: convolutional neural network.}
     \label{ch7_fig:CMs}
\end{figure}

The results presented in Table \ref{ch7_tab:main_test_set} show that the two deep learning-based approaches which utilise the silver-standard PVS masks---either directly (the multi-task CNN), or via the segmentation U-Net's pseudo-label predictions (the conditional U-Net)---strongly out perform both the baseline CNN, which does not utilise any segmentations, and the logistic regression models, which utilises only basic features derived from the segmentation U-Net's pseudo-labels, in addition to age (BG model) and sex (CSO model). The multi-task learning approach exhibits the best or joint best performance across all metrics, and is the only method to exhibit a statistically significant ($p=0.03$) improvement over the logistic regression models on mean F1 score, according to paired difference bootstrap testing (pairwise statistics in \ref{ch7_app:stats}). Confusion matrices are shown in Figure \ref{ch7_fig:CMs}.

\begin{table}[htbp]
\centering
\caption{\textbf{Comparative results on classification metrics}. Comparison of methods for perivascular space (PVS) score prediction on the test set. Reported as mean {[}95\% CIs{]} with confidence intervals (CIs) calculated over 10,000 bootstrap iterations with best results for each metric in bold. mAP: mean average precision, F1: macro-average F1 score, ACC: accuracy, BG: basal ganglia, CSO: centrum semiovale, CNN: convolutional neural network.}
\label{ch7_tab:main_test_set}
\resizebox{\textwidth}{!}{%
\begin{tabular}{l|ccc|ccc|ccc}
\hline
\multirow{2}{*}{Method} &
  \multicolumn{3}{c|}{Mean} &
  \multicolumn{3}{c|}{BG} &
  \multicolumn{3}{c}{CSO} \\ \cline{2-10} 
 &
  mAP &
  F1 &
  ACC &
  mAP &
  F1 &
  ACC &
  mAP &
  F1 &
  ACC \\ \hline
Logistic regression &
  \begin{tabular}[c]{@{}c@{}}49.32\\ {[}45.65, 55.18{]}\end{tabular} &
  \begin{tabular}[c]{@{}c@{}}49.58\\ {[}44.74, 54.10{]}\end{tabular} &
  \begin{tabular}[c]{@{}c@{}}52.62\\ {[}48.19, 57.26{]}\end{tabular} &
  \begin{tabular}[c]{@{}c@{}}58.19\\ {[}51.80, 66.59{]}\end{tabular} &
  \begin{tabular}[c]{@{}c@{}}57.06\\ {[}49.90, 63.74{]}\end{tabular} &
  \begin{tabular}[c]{@{}c@{}}61.69\\ {[}55.65, 67.74{]}\end{tabular} &
  \begin{tabular}[c]{@{}c@{}}40.45\\ {[}36.77, 46.71{]}\end{tabular} &
  \begin{tabular}[c]{@{}c@{}}42.09\\ {[}35.75, 48.17{]}\end{tabular} &
  \begin{tabular}[c]{@{}c@{}}43.55\\ {[}37.50, 49.60{]}\end{tabular} \\
Baseline CNN &
  \begin{tabular}[c]{@{}c@{}}52.11\\ {[}48.24, 58.27{]}\end{tabular} &
  \begin{tabular}[c]{@{}c@{}}51.95\\ {[}46.93, 56.64{]}\end{tabular} &
  \begin{tabular}[c]{@{}c@{}}53.23\\ {[}48.39, 57.86{]}\end{tabular} &
  \begin{tabular}[c]{@{}c@{}}58.08\\ {[}51.86, 66.05{]}\end{tabular} &
  \begin{tabular}[c]{@{}c@{}}57.11\\ {[}50.28, 63.33{]}\end{tabular} &
  \begin{tabular}[c]{@{}c@{}}58.47\\ {[}52.42, 64.52{]}\end{tabular} &
  \begin{tabular}[c]{@{}c@{}}46.15\\ {[}41.70, 53.55{]}\end{tabular} &
  \begin{tabular}[c]{@{}c@{}}46.80\\ {[}40.27, 52.93{]}\end{tabular} &
  \begin{tabular}[c]{@{}c@{}}47.98\\ {[}41.94, 54.03{]}\end{tabular} \\
Conditional CNN &
  \begin{tabular}[c]{@{}c@{}}60.22\\ {[}55.91, 66.10{]}\end{tabular} &
  \begin{tabular}[c]{@{}c@{}}55.11\\ {[}50.12, 59.78{]}\end{tabular} &
  \begin{tabular}[c]{@{}c@{}}57.06\\ {[}52.42, 61.69{]}\end{tabular} &
  \begin{tabular}[c]{@{}c@{}}64.92\\ {[}58.14, 73.66{]}\end{tabular} &
  \begin{tabular}[c]{@{}c@{}}58.85\\ {[}51.95, 65.30{]}\end{tabular} &
  \begin{tabular}[c]{@{}c@{}}62.90\\ {[}56.85, 68.95{]}\end{tabular} &
  \begin{tabular}[c]{@{}c@{}}55.52\\ {[}49.89, 62.48{]}\end{tabular} &
  \begin{tabular}[c]{@{}c@{}}51.37\\ {[}45.05, 57.39{]}\end{tabular} &
  \textbf{\begin{tabular}[c]{@{}c@{}}51.21\\ {[}45.16, 57.26{]}\end{tabular}} \\
Multi-task CNN &
  \textbf{\begin{tabular}[c]{@{}c@{}}64.08\\ {[}59.74, 69.27{]}\end{tabular}} &
  \textbf{\begin{tabular}[c]{@{}c@{}}57.23\\ {[}52.49, 61.46{]}\end{tabular}} &
  \textbf{\begin{tabular}[c]{@{}c@{}}58.47\\ {[}54.03, 62.70{]}\end{tabular}} &
  \textbf{\begin{tabular}[c]{@{}c@{}}69.89\\ {[}63.34, 76.69{]}\end{tabular}} &
  \textbf{\begin{tabular}[c]{@{}c@{}}62.95\\ {[}56.38, 69.11{]}\end{tabular}} &
  \textbf{\begin{tabular}[c]{@{}c@{}}65.73\\ {[}59.68, 71.37{]}\end{tabular}} &
  \textbf{\begin{tabular}[c]{@{}c@{}}58.27\\ {[}52.35, 65.36{]}\end{tabular}} &
  \textbf{\begin{tabular}[c]{@{}c@{}}51.51\\ {[}45.00, 57.42{]}\end{tabular}} &
  \textbf{\begin{tabular}[c]{@{}c@{}}51.21\\ {[}44.76, 57.26{]}\end{tabular}} \\ \hline
\end{tabular}%
}
\end{table}

\subsection{Multi-task model validation}

In this section, we performed additional experiments to validate our best-performing model: the multi-task CNN. Given the difficulty of accurately predicting the PVS score with even our best model, we consider four avenues for evaluating its real-world utility. The first is through the use of 3D gradient-weighted class activation mapping (GradCAM) \citep{selvaraju2020grad}, to confirm that the model actually uses the PVS visible in the image to guide its prediction. The second is via visualisation of the distribution of probabilistic outputs of the model. The third is an evaluation of clinical validity: assessing whether the model's predicted PVS scores share the same associations with established clinical biomarkers as the trained observer's ground truth scores. The fourth is an evaluation of inter-rater reliability comparing our model and two trained observers.

\begin{figure}[htbp]
     \centering
     \begin{subfigure}[b]{0.32\textwidth}
         \centering
         \includegraphics[width=\textwidth]{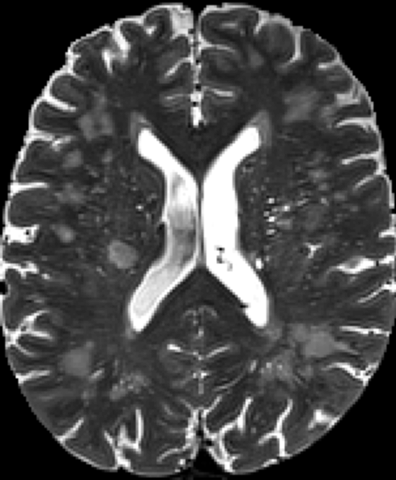}
     \end{subfigure}
     \hfill
     \begin{subfigure}[b]{0.32\textwidth}
         \centering
         \includegraphics[width=\textwidth]{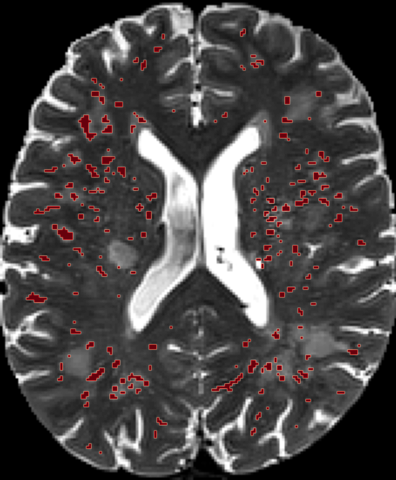}
     \end{subfigure}
     \hfill
     \begin{subfigure}[b]{0.32\textwidth}
         \centering
         \includegraphics[width=\textwidth]{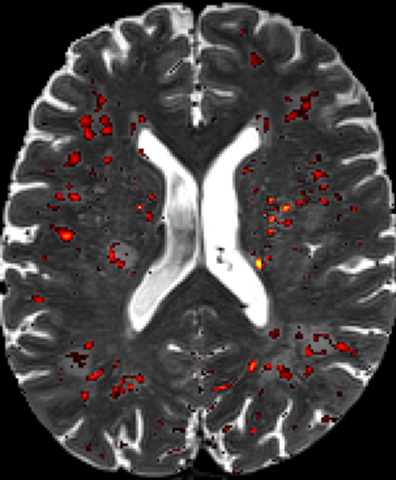}
     \end{subfigure}
     \caption{\textbf{The multi-task CNN localises PVS}. \textbf{Left}: T2-weighted (T2w) magnetic resonance imaging (MRI) scan with a perivascular space (PVS) score of 3 for the basal ganglia (BG) and centrum semiovale (CSO). \textbf{Middle}: Silver-standard PVS mask. \textbf{Right}: Gradient-weighted class activation map (GradCAM) for the first layer of the multi-task CNN (random seed 2), using the mean of the predicted probability of score 3 for BG and CSO as the target for calculation of the gradient. We can see that some of the highest areas of activation are on visible PVS. CNN: convolutional neural network.}
     \label{ch7_fig:cam0_main}
\end{figure}

Figure \ref{ch7_fig:cam0_main} shows the GradCAM for the first layer of the multi-task CNN on an image with a high PVS burden. While care should be taken not to over-interpret visualisations of this kind, we can, very encouragingly, see that several of the most salient features are PVS. This provides us with evidence that our model is indeed performing classification of the PVS score based (at least partially) on the actual PVS, and not based solely on other correlated feature of the image. An extended GradCAM-based analysis can be found in \ref{ch7_app:cam}, where we show that the baseline and conditional CNNs' maps to not appear to localise PVS in the first layer at all.

\begin{figure}[htbp]
     \centering
     \begin{subfigure}[b]{0.48\textwidth}
         \centering
         \includegraphics[width=\textwidth]{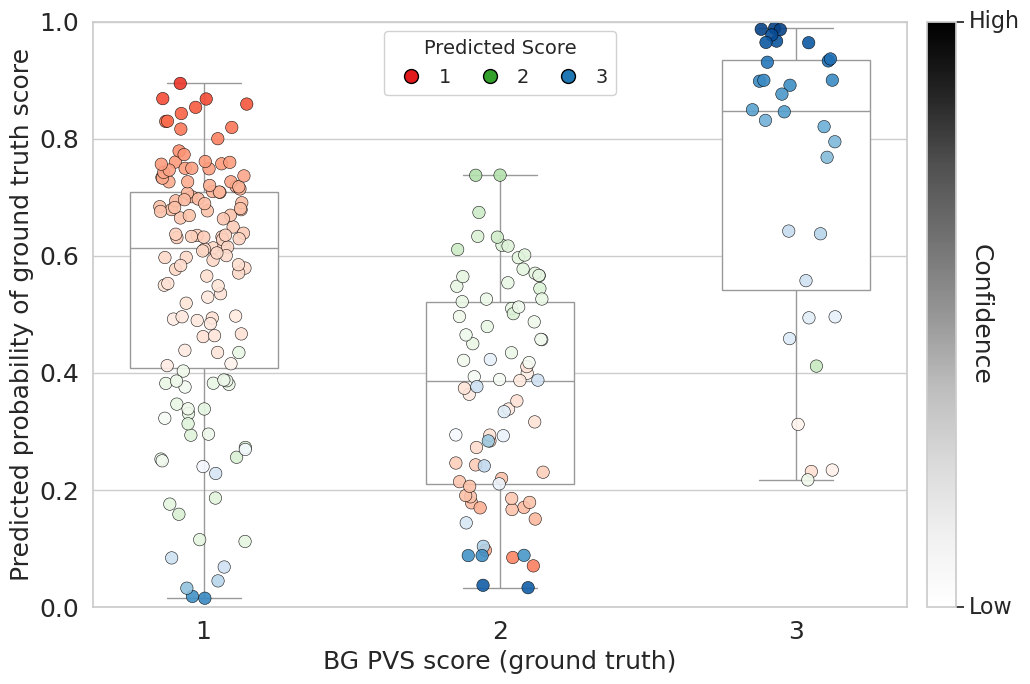}
     \end{subfigure}
     \hfill
     \begin{subfigure}[b]{0.48\textwidth}
         \centering
         \includegraphics[width=\textwidth]{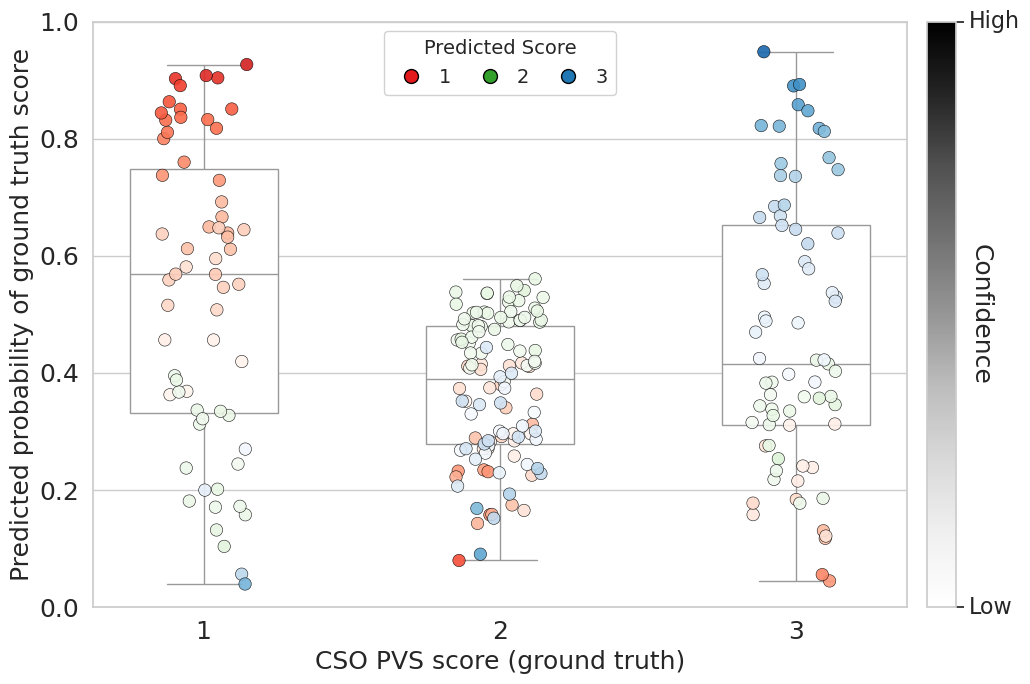}
     \end{subfigure}
     \hfill
     \begin{subfigure}[b]{0.48\textwidth}
         \centering
         \includegraphics[width=\textwidth]{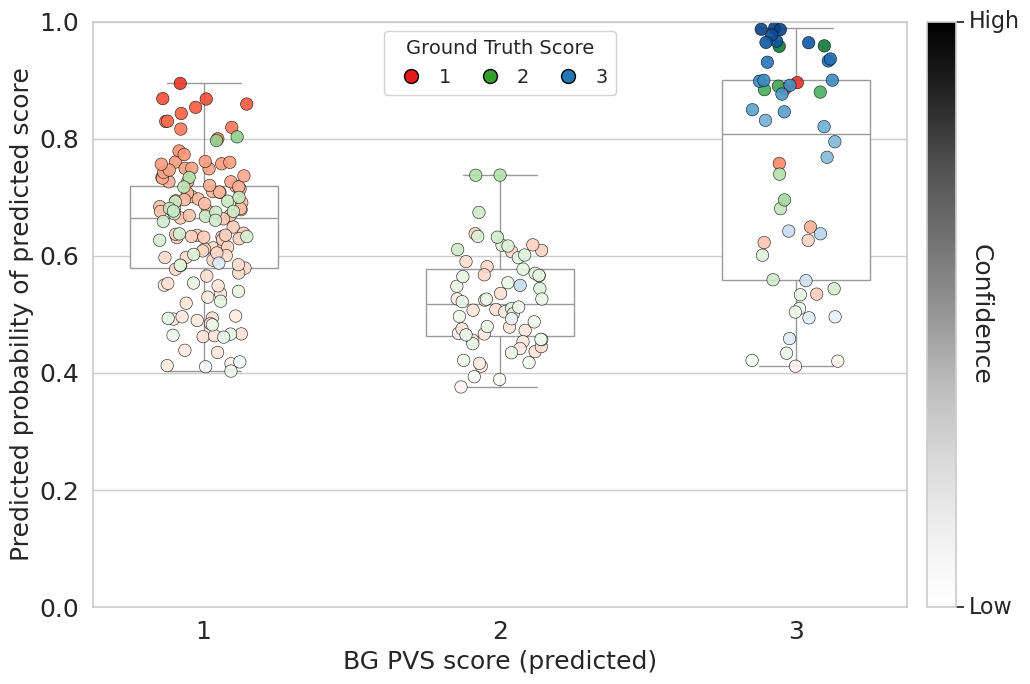}
     \end{subfigure}
     \hfill
     \begin{subfigure}[b]{0.48\textwidth}
         \centering
         \includegraphics[width=\textwidth]{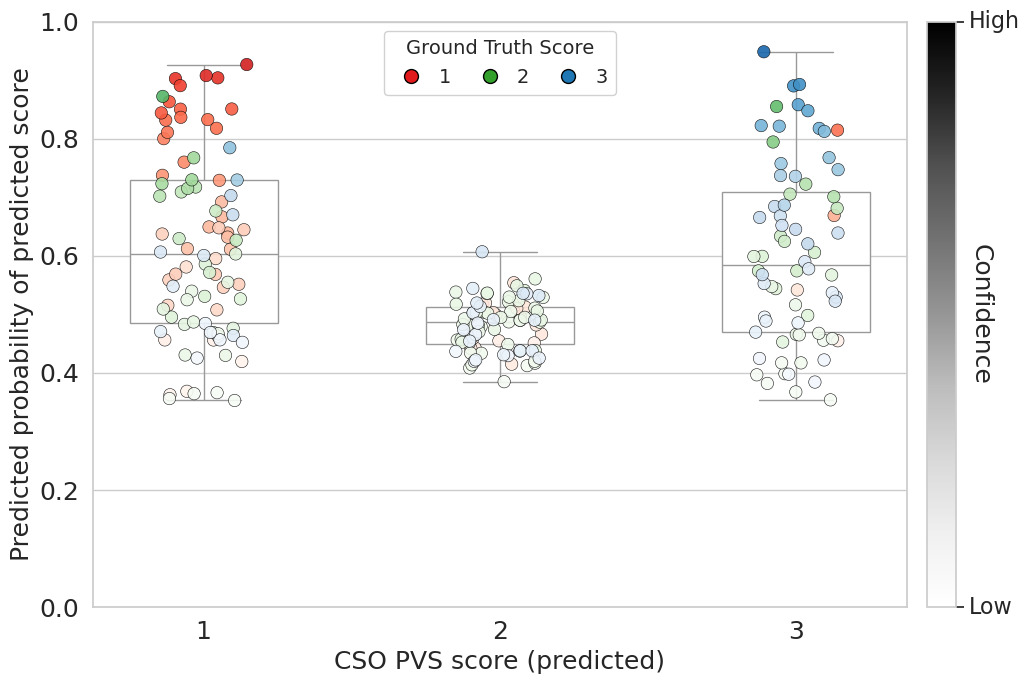}
     \end{subfigure}
     \caption{\textbf{Distribution and confidence of the multi-task CNN's PVS score predictions}. Boxplots showing the distribution of predicted probabilities from the multi-task CNN for each ground truth label (top) and predicted label (bottom). Confidence measured as the maximum possible Shannon's entropy minus the actual Shannon's Entropy, given by $-\log_2(\frac{1}{3}) + \sum_{s=1}^3 p_s \log_2 (p_s)$ for predicted probability $p_s$ of score $s$. CNN: convolutional neural network, PVS: perivascular space.}
     \label{ch7_fig:ent}
\end{figure}

Figure \ref{ch7_fig:ent} shows the distribution of predicted probabilities for each ground truth class (top) and predicted class (bottom), and can be used to inspect the confidence of true positive (both, correct colour), false positive (bottom, wrong colour), and false negative (top, wrong colour) predictions. We can see from the bottom row, for both BG and CSO PVS, the model is less confident in predicting scores of 2, than 1 or 3. This is a reassuring finding. In our modified scale, a score of 2 represents a specific intermediate count (11-20 individual PVS). In contrast, scores of 1 include subjects with $\leq$10 PVS (including perfectly healthy subjects) and scores of 3 represent an open-ended interval with $\geq$21 PVS. Because a score of 2 is bounded on both sides, the margin for error is smaller, and the sensitivity required to distinguish these cases is higher. Low confidence on this difficult intermediate class---even when the prediction is correct---suggests that the model is evaluating true PVS-relevant features rather than relying on spurious correlations.

From the bottom row we can see that in general there is a correspondence between high confidence predictions and accuracy. With the exception of CSO PVS scores of 2, there are noticeably more correct predictions than incorrect predictions above the 75th percentile of predicted probabilities. Additionally, high confidence false positives for scores 1 and 3 are both more frequently false negatives for score 2, than for scores 3 or 1, providing further evidence to suggest that the model has discovered the ordinal nature of the data.

Next, we utilised a subset of the test set (200/248) with a number of clinical variables available: age, sex, hypertension, white matter hyperintensity (WMH) volume, and presence of any ischaemic stroke lesions (ISL). We fit two ordinal logistic regression models on this data; one model used the trained observer's ground truth PVS scores as the dependent variable, while the other model used the predicted PVS scores from the multi-task CNN as the dependent variable.

Examining the results displayed in Table \ref{ch7_tab:betas}, we can see that to optimally model both the ground truth scores and the predicted scores using these clinical variables, similar normalised beta coefficients are needed. All betas are in agreement regarding the direction of the correlation and the most statistically significant predictor is white matter hyperintensity volume for all models. From a clinical perspective, these results are encouraging. They demonstrate that the multi-task CNN predictions preserve the expected relationships with established cerebrovascular risk factors.
 
\begin{table}[htbp]
\centering
\caption{\textbf{Association of clinical variables with the multi-task CNN's PVS score predictions}. Normalised beta coefficients and p-values of predictor variables when either the ground truth or multi-task CNN predicted perivascular space (PVS) scores are the target variables of an ordinal logistic regression model. Reported on a subset (N=200) of the test set with available predictor variables. BG: basal ganglia, CSO: centrum semiovale, WMH: white matter hyperintensity, ISL: ischaemic stroke lesion, cont: continuous, bin: binary, CNN: convolutional neural network.}
\label{ch7_tab:betas}
\resizebox{\textwidth}{!}{%
\begin{tabular}{l|ll|ll}
\hline
\multirow{2}{*}{Predictor} & \multicolumn{2}{c|}{BG}             & \multicolumn{2}{c}{CSO}             \\ \cline{2-5} 
                   & \multicolumn{1}{c}{Ground truth} & \multicolumn{1}{c|}{Predicted} & \multicolumn{1}{c}{Ground truth} & \multicolumn{1}{c}{Predicted} \\ \hline
Age (cont.)                & -0.0689 (0.6498) & -0.0683 (0.6623) & 0.0117 (0.9355)  & 0.0296 (0.8399)  \\
Female (bin.)              & -0.3360 (0.2614) & -0.0921 (0.7618) & 0.4463 (0.1070)  & 0.2618 (0.3416)  \\
Hypertension (bin.)        & 0.5308 (0.0844)  & 0.5477 (0.0847)  & 0.3699 (0.2034)  & 0.3601 (0.2089)  \\
WMH volume (cont.) & 0.7276 (\textless{}0.0001)*      & 1.2834 (\textless{}0.0001)*    & 0.2716 (0.0546)                  & 0.7027 (0.0007)*              \\
ISL (bin.)                 & 0.5916 (0.0729)  & 0.5336 (0.1182)  & -0.0438 (0.8897) & -0.0712 (0.8259) \\ \hline
\end{tabular}%
}
\end{table}

\begin{figure}[htbp]
     \centering
     \begin{subfigure}[b]{0.49\textwidth}
         \centering
         \includegraphics[width=\textwidth]{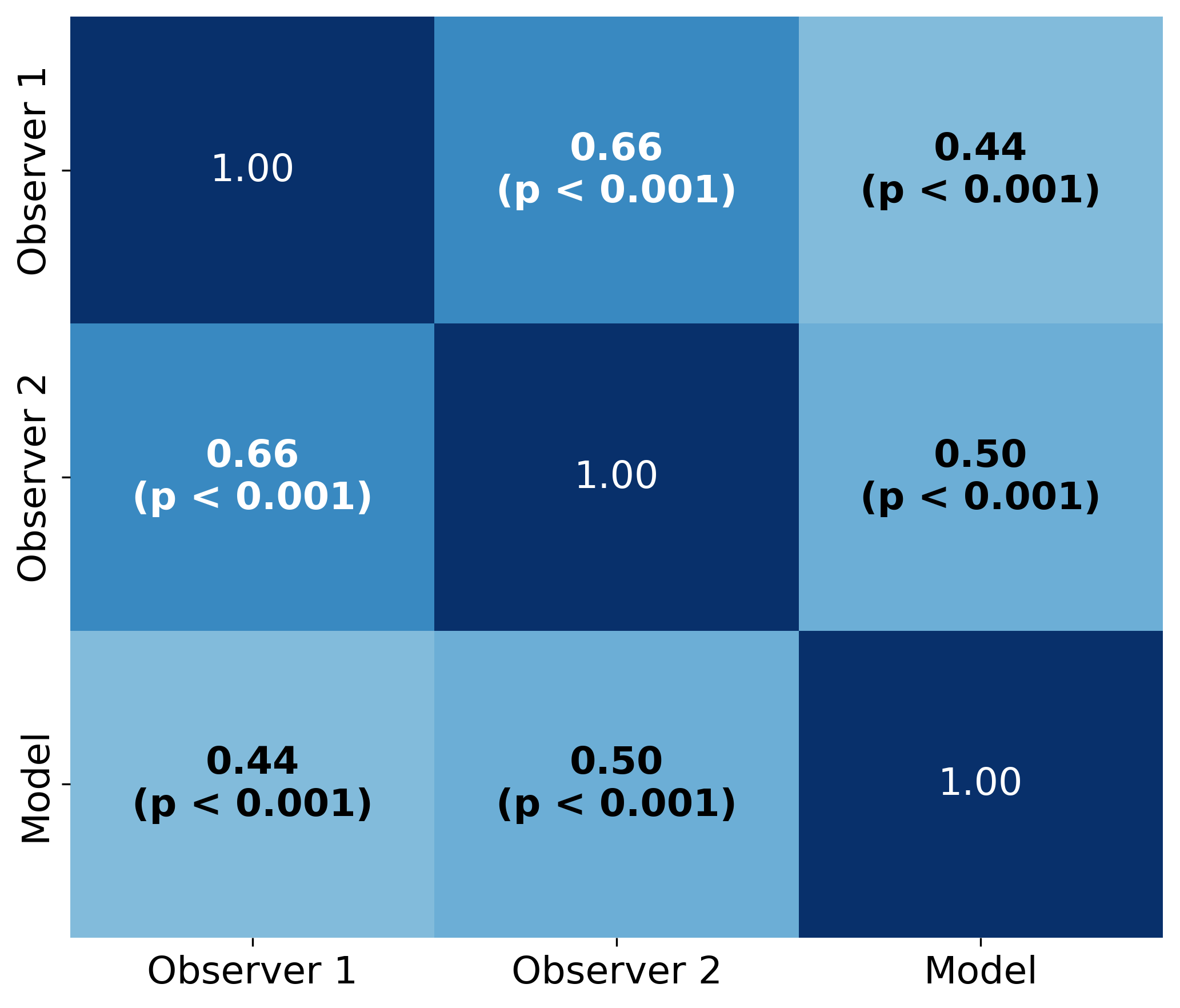}
         \caption{BG PVS score}
     \end{subfigure}
     \hfill
     \begin{subfigure}[b]{0.49\textwidth}
         \centering
         \includegraphics[width=\textwidth]{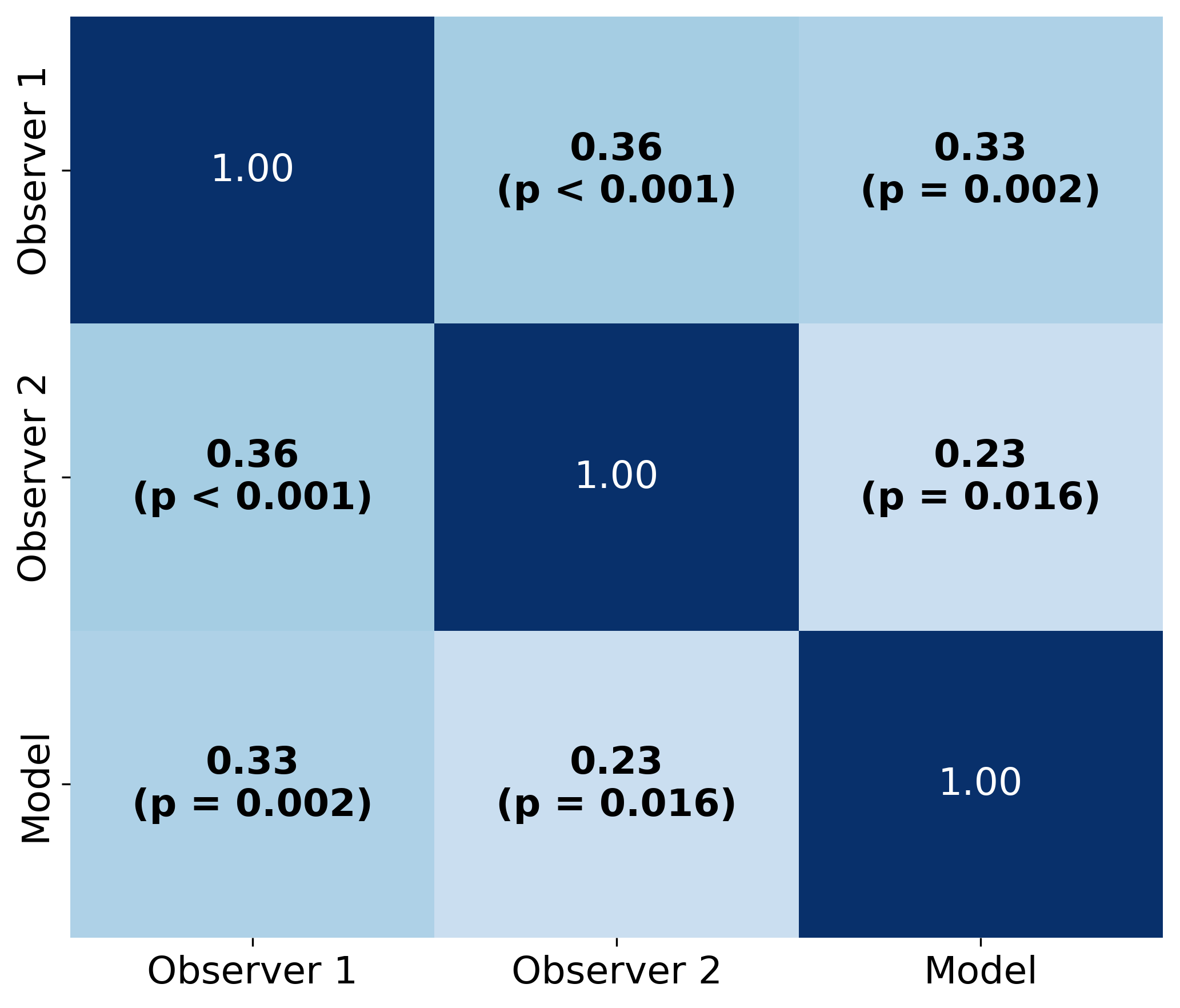}
         \caption{CSO PVS score}
     \end{subfigure}
     \caption{\textbf{Inter-rater reliability}. Linear-weighted Cohen's Kappa between two observers and the multi-task CNN model. Reported on a subset (N=62) of the test rated by two independent trained observers. P-values corrected for multiple comparisons (Bonferroni). CNN: convolutional neural network, PVS: perivascular space.}
     \label{ch7_fig:inter-rater}
\end{figure}

Finally, for a small subset of the test set (62/248), two trained observers independently provided PVS scores. Figure \ref{ch7_fig:inter-rater} shows the linear-weighted Cohen's Kappa between the two trained observers and the multi-task CNN model. On BG PVS score, the two trained observers have the highest agreement, and the CNN model has higher agreement with observer 2 than observer 1. On CSO PVS score, the trained observers again have the highest agreement, but it is considerably lower than their agreement on BG PVS scores (a result also observed in \cite{potter2015cerebral}) and also lower than the observer-model agreement on BG PVS scores. The model has higher agreement with observer 1 than observer 2 for CSO PVS scores. While the level of agreement between observers and between the model and observers are quite similar, these results---although limited in their power by the size of the dataset---indicate that even for this difficult and (in practice) subjective task, human observers are likely still more reliable than our automated multi-task CNN, when reliability is measured between raters. An analysis of intra-rater variability would, of course, highlight the algorithmic consistency inherent to automated computational methods, which is a highly desirable trait.

\section{Conclusion}

In this work, we developed and evaluated three distinct modelling methodologies for utilising silver-standard PVS segmentation masks to predict PVS scores on a modified Potters/Wardlaw scale. We demonstrated that integrating this segmentation data enables the training of a more accurate CNN-based model, with multi-task learning emerging as the most successful paradigm. Our GramCAM-based analysis revealed that multi-task learning resulted in a model which learned to localise individual PVS---a quality not shown by the baseline or conditional CNNs.

The evaluation of this multi-task model highlighted its clinical utility. Although its inter-rater agreement with independent trained observers was lower than the agreement between the observers themselves, the inter-rater agreement of the observers was only 0.66 and 0.36 for the BG and CSO PVS scores respectively, highlighting the difficulty of the task and the need for standardisation via automation. The model behaved in a probabilistically sensible manner, exhibiting appropriately lower confidence on more difficult classes. Furthermore, the model’s predictions maintained similar associations with established clinical markers of neurovascular health as the trained observer’s ground truth, demonstrating strong clinical validity.

Currently, our model utilises a modified version of the Potters/Wardlaw scale due to the limited number of subjects presenting with scores of 0 or 4 in our datasets. To maximise the clinical applicability of this approach, future work should focus on reproducing this model using the full Potters/Wardlaw scale once sufficient training data is aggregated. Ideally, this expansion should also incorporate the binary midbrain PVS score.

Ultimately, automating PVS scoring addresses an immediate need in neurovascular research. While accurate automated segmentation represents the long-term ideal for deriving repeatable, quantitative measures, visual scoring remains the current clinical standard because it is fast, practical, and widely applicable. By learning to replicate the implicit policy of a trained observer, our automated scoring model bridges this gap, providing a tool for identifying health associations and improving disease risk modelling today, while the field continues to advance toward robust segmentation in the future.

\paragraph{Limitations}
Although we attempted to account for the variability in results on the test data via bootstrapping, and the variability in the training process via ensembling the results of multiple independent training runs, the use of a single data split is a limitation of this study.

\section*{Acknowledgements}

J.P. is funded by Medical Research Scotland [ref. PHD-50441-2021] and Canon Medical Research Europe. Funding from Row Fogo Charitable Trust (Ref No: AD.ROW4.35. BRO-D.FID3668413), and the UK Medical Research Council (UK Dementia Research Institute at the University of Edinburgh, award number UK DRI-4002; G0700704/84698) are also gratefully acknowledged. M.O.B. gratefully acknowledges funding from: EPSRC grant no. EP/X025705/1; British Heart Foundation and The Alan Turing Institute Cardiovascular Data Science Award (C-10180357); the SCONe projects funded by Chief Scientist Office, Edinburgh \& Lothians Health Foundation, Sight Scotland, the Royal College of Surgeons of Edinburgh, the RS Macdonald Charitable Trust, and Fight For Sight.


The original studies which produced the data used in this paper were supported by: the UK Dementia Research Institute which receives its funding from DRI Ltd, funded by the UK MRC, Alzheimer’s Society and Alzheimer’s Research UK; the Fondation Leducq Network for the Study of Perivascular Spaces in Small Vessel Disease (16 CVD 05); Stroke Association ``Small Vessel Disease-Spotlight on Symptoms (SVD-SOS)''(SAPG 19\textbackslash100068); The Row Fogo Charitable Trust Centre for Research into Aging and the Brain; Stroke Association Garfield Weston Foundation Senior Clinical Lectureship (FND) (TSALECT 2015/04); NHS Research Scotland (FND); Stroke Association Post-Doctoral Fellowship (SW) (SAPDF 18/100026); British Heart Foundation Edinburgh Centre for Research Excellence (RE/18/5/34216); British Heart Foundation (SP/07/001/23603; PG/08/103; PG/12/29/29497; CS/13/1/30327), NHS Lothian Research and Development Office (MJT); European Union Horizon 2020, PHC-03–15, project No666881, ``SVDs@Target'' (MS,GB); Chief Scientist Office of Scotland Clinical Academic Fellowship (UC) (CAF/18/08); Stroke Association Princess Margaret Research Development Fellowship (UC) (2018); Medical Research Scotland studentship (AM) (PhD-1165–2017); the Wellcome Trust and the Royal Society (WT088134/Z/09/A, 082464/Z/07/Z, 221890/Z/20/Z); College of Medicine and Veterinary Medicine, University of Edinburgh scholarship, as part of the Wellcome-funded Translational Neuroscience PhD programme (OH); MRC Doctoral Training Programme in Precision Medicine (CM) (MR/R01566X/1); Age-UK disconnected mind study; The Brain Research Imaging Centre Edinburgh; the Economic and Social Research Council; and the Milton Damerel Trust. The Research MR scanners are supported by the Scottish Funding Council through the Scottish Imaging Network, A Platform for Scientific Excellence (SINAPSE) Collaboration; the 3T scanner is funded by the Wellcome Trust (104916/Z/14/Z), Dunhill Trust (R380R/1114), Edinburgh and Lothians Health Foundation (2012/17), The Medical Research Council (G0701120; G1001245), Biotechnology and Biological Sciences Research Council, Muir Maxwell Research Fund and the University of Edinburgh.

\clearpage

\appendix

\section{Model configuration details} \label{ch7_app:configs}

\subsection{Common data augmentation and preprocessing}

For all deep learning models, data augmentation was implemented with TorchIO \citep{perez2021torchio} and consisted of the following random operations: isotropic affine transformation ($p=1.0$, scale=[0.9, 1.1], rotation=[-90, +90]); Gaussian noise ($p=0.15$, mean=0, std=[0, 0.316]); Gaussian blur ($p=0.1$, std=[0.5, 1.5]); brightness adjustment ($p=0.15$, multiplier=[0.7, 1.3]); contrast adjustment ($p=0.15$, gamma=[0.65, 1.5]); flip along any axis ($p=0.5$); and low resolution simulation in any axis via nearest neighbour downsampling and linear upsampling ($p=0.25$, factor=[1, 2]).

All models were trained using full 3D image patches. Before being passed into the network, the augmented patches are renormalised so voxel intensities are in the range [0, 1] using the brain mask, saturating at the 2nd and 98th percentiles.

\subsection{Segmentation U-Net}

We trained the U-Net segmentation model using best practices developed in nnU-Net \citep{isensee2021nnu}. The DynUNet model from MONAI \citep{cardoso2022monai} was used to instantiate the architecture, with 6 levels of spatial resolution with 32, 64, 128, 256, 320, and 320 channels respectively, $3 \times 3 \times 3$ kernels, residual blocks, instance normalisation, Leaky ReLU activations with a slope coefficient of 0.01, and three levels of additional deep supervision, with normalised exponentially decreasing loss weights (as per nnU-Net).

The segmentation loss function was the equally weighted sum of Dice and Cross-Entropy loss, using stochastic gradient descent and polynomial learning rate decay starting at 0.01 with a power of 0.9, as per nnU-Net. We employed an effective batch size of 6, achieved via gradient accumulation. The model was trained for 1000 epochs.

\subsection{Baseline and conditional CNNs}

We used a 3D ResNet model from MONAI with 1, 2, 2, and 4 blocks per layer and with 32, 64, 128, and 256 channels, respectively. To preserve small features like PVS, we omitted the initial max pooling layer. We also applied a $1 \times 1 \times 1$ convolution and a $3 \times 3 \times 3$ convolution in parallel to the standard initial $7 \times 7 \times 7$ convolution, summing the results prior to the first normalisation and activation. 

We replaced the single prediction head with two heads, one for the BG and one for the CSO. The Cross-Entropy loss was used as the classification loss function, with class weights selected to balance the dataset and a label smoothing factor of 0.1. We used group normalisation with 8 groups, ReLU activation, and the AdamW optimiser (learning rate = 0.001, betas = (0.9, 0.999)). We employed an effective batch size of 220, achieved via gradient accumulation. In addition to the three MRI sequences, BG and CSO masks were also input to the network. The conditional CNN additionally included the PVS pseudo-labels produced by the segmentation U-Net as an input channel. The models were trained for 1000 epochs.

\subsection{Multitask CNN}

We used the same 3D ResNet architecture as above, but added a lightweight segmentation decoder. The decoder is symmetric with the encoder, but with only a single ResNet block per layer. The lower-resolution feature maps of the previous layer are upsampled with transposed convolutions and concatenated with skip connections from the equivalent encoder layer output (as in a standard U-Net) before each layer. 

For the loss function, we set the base task weighting $\alpha=0.5$ (refer to the main paper), which is dynamically scaled by the parameter $\beta$ to appropriately balance the segmentation and classification components (refer to the main paper). All other training details are identical to the baseline CNN.

\subsection{Logistic regression}

The details of our multinomial logistic regression model selection approach are covered in the main body of this paper. The feature pool included age, sex, and the following PVS statistics derived from the segmentation U-Net's pseudo-labels: total volume and count (calculated both globally and masked to the BG and CSO), and the maximum single-slice volume and count per hemisphere (calculated only within the BG and CSO).

All features (except sex) were z-score standardised. The standardisation statistics (mean and standard deviation) were calculated exclusively on the training set and then applied to the validation and test sets. The LBFGS optimiser with a maximum of 5000 iterations and an L2 penalty was used.

\clearpage
\section{Silver-standard masks vs pseudo-labels} \label{ch7_app:abl}

We restricted the training, validation and test sets to the subset containing silver-standard PVS masks, and evaluated the impact of using the pseudo-labels produced by our segmentation U-Net instead, for training the logistic regression and conditional CNN models.

\subsection{Logistic regression}

We re-performed feature and kernel selection for the logistic regression models, using the data subset. 

For the BG model using the original silver-standard PVS masks a degree 2 polynomial with four features ((1) age; (2) PVS volume in the whole scan; (3) PVS volume in the BG; and (4) the maximum count of BG PVS in one hemisphere of any slice) was selected, and for the BG model using the pseudo-labels a degree 2 polynomial with two features ((1) PVS volume in the BG; and (2) the maximum count of BG PVS in one hemisphere of any slice) selected. 

For the CSO model using the original silver-standard PVS masks a degree 1 polynomial with five features ((1) sex; (2) age; (3) PVS volume in the whole scan; (4) PVS volume in the CSO; and (5) PVS count in the CSO) was selected, and for the CSO model using the pseudo-labels a degree 3 polynomial with five features ((1) sex; (2) age; (3) PVS volume in the whole scan; (4) PVS volume in the CSO; and (5) the maximum count of CSO PVS in one hemisphere of any slice) selected.

Table \ref{ch7_abl:lr_tab} shows the difference in results on the test subset. There are no statistically significant differences on any metrics (the pairwise comparison p-values are presented in \ref{ch7_app:stats}). 

\begin{table}[htbp]
\centering
\caption{\textbf{Ablation of PVS mask sources --- logistic regression}. Logistic regression model performance when features are derived from the original silver-standard perivascular space (PVS) masks, or the binarised pseudo-labels produced by the segmentation U-Net. Reported as mean {[}95\% CIs{]} with confidence intervals (CIs) calculated over 10,000 bootstrap iterations with best results for each metric in bold. Reported on a subset (N=197) of the test data where silver-standard PVS masks are available. mAP: mean average precision, F1: F1 score, ACC: accuracy, BG: basal ganglia, CSO: centrum semiovale.}
\label{ch7_abl:lr_tab}
\resizebox{\textwidth}{!}{%
\begin{tabular}{l|ccc|ccc|ccc}
\hline
\multirow{2}{*}{Mask type} & \multicolumn{3}{c|}{Mean} & \multicolumn{3}{c|}{BG} & \multicolumn{3}{c}{CSO} \\ \cline{2-10} 
                           & mAP     & F1     & ACC    & mAP    & F1    & ACC    & mAP    & F1    & ACC    \\ \hline
Original &
  \begin{tabular}[c]{@{}c@{}}53.06\\ {[}48.39, 59.76{]}\end{tabular} &
  \textbf{\begin{tabular}[c]{@{}c@{}}52.67\\ {[}46.99, 57.79{]}\end{tabular}} &
  \begin{tabular}[c]{@{}c@{}}54.57\\ {[}49.24, 59.64{]}\end{tabular} &
  \begin{tabular}[c]{@{}c@{}}58.13\\ {[}51.32, 67.42{]}\end{tabular} &
  \textbf{\begin{tabular}[c]{@{}c@{}}61.82\\ {[}53.79, 68.92{]}\end{tabular}} &
  \textbf{\begin{tabular}[c]{@{}c@{}}65.48\\ {[}58.88, 72.08{]}\end{tabular}} &
  \begin{tabular}[c]{@{}c@{}}48.00\\ {[}42.71, 55.53{]}\end{tabular} &
  \begin{tabular}[c]{@{}c@{}}43.52\\ {[}36.56, 50.27{]}\end{tabular} &
  \begin{tabular}[c]{@{}c@{}}43.65\\ {[}37.06, 50.76{]}\end{tabular} \\
Pseudo-labels &
  \textbf{\begin{tabular}[c]{@{}c@{}}54.00\\ {[}49.46, 60.46{]}\end{tabular}} &
  \begin{tabular}[c]{@{}c@{}}51.35\\ {[}45.50, 56.71{]}\end{tabular} &
  \textbf{\begin{tabular}[c]{@{}c@{}}55.84\\ {[}50.76, 61.17{]}\end{tabular}} &
  \textbf{\begin{tabular}[c]{@{}c@{}}58.58\\ {[}51.69, 67.44{]}{]}\end{tabular}} &
  \begin{tabular}[c]{@{}c@{}}56.24\\ {[}47.82, 63.90{]}\end{tabular} &
  \begin{tabular}[c]{@{}c@{}}64.97\\ {[}58.38, 71.57{]}\end{tabular} &
  \textbf{\begin{tabular}[c]{@{}c@{}}49.43\\ {[}43.93, 57.23{]}\end{tabular}} &
  \textbf{\begin{tabular}[c]{@{}c@{}}46.46\\ {[}39.14, 53.49{]}\end{tabular}} &
  \textbf{\begin{tabular}[c]{@{}c@{}}46.70\\ {[}39.59, 53.81{]}\end{tabular}} \\ \hline
\end{tabular}%
}
\end{table}

\subsection{Conditional CNN}

Table \ref{ch7_abl:conditional_tab} shows the difference in results on the test subset. There are no statistically significant differences on any metrics (the pairwise comparison p-values are presented in \ref{ch7_app:stats}). For this ablation, we only trained one random seed for each model. 

\begin{table}[htbp]
\centering
\caption{\textbf{Ablation of PVS mask sources --- conditional CNN}. Conditional CNN model performance when input masks are the original silver-standard perivascular space (PVS) masks,  or the binarised pseudo-labels produced by the segmentation U-Net. Reported as mean {[}95\% CIs{]} with confidence intervals (CIs) calculated over 10,000 bootstrap iterations with best results for each metric in bold. Reported on a subset (N=197) of the test data where silver-standard PVS masks are available. mAP: mean average precision, F1: F1 score, ACC: accuracy, BG: basal ganglia, CSO: centrum semiovale.}
\label{ch7_abl:conditional_tab}
\resizebox{\textwidth}{!}{%
\begin{tabular}{l|ccc|ccc|ccc}
\hline
\multirow{2}{*}{Mask type} & \multicolumn{3}{c|}{Mean} & \multicolumn{3}{c|}{BG} & \multicolumn{3}{c}{CSO} \\ \cline{2-10} 
                           & mAP     & F1     & ACC    & mAP    & F1    & ACC    & mAP    & F1    & ACC    \\ \hline
Original &
  \begin{tabular}[c]{@{}c@{}}55.07\\ {[}50.49, 61.75{]}\end{tabular} &
  \begin{tabular}[c]{@{}c@{}}51.40\\ {[}45.63, 56.71{]}\end{tabular} &
  \begin{tabular}[c]{@{}c@{}}52.03\\ {[}46.70, 57.36{]}\end{tabular} &
  \textbf{\begin{tabular}[c]{@{}c@{}}61.23\\ {[}53.86, 70.57{]}\end{tabular}} &
  \begin{tabular}[c]{@{}c@{}}55.81\\ {[}48.16, 62.91{]}\end{tabular} &
  \begin{tabular}[c]{@{}c@{}}56.35\\ {[}49.24, 63.45{]}\end{tabular} &
  \begin{tabular}[c]{@{}c@{}}48.91\\ {[}43.22, 57.05{]}\end{tabular} &
  \begin{tabular}[c]{@{}c@{}}46.99\\ {[}39.56, 53.95{]}\end{tabular} &
  \begin{tabular}[c]{@{}c@{}}47.72\\ {[}40.61, 54.82{]}\end{tabular} \\
Pseudo-labels &
  \textbf{\begin{tabular}[c]{@{}c@{}}57.01\\ {[}52.28, 64.07{]}\end{tabular}} &
  \textbf{\begin{tabular}[c]{@{}c@{}}53.30\\ {[}47.67, 58.55{]}\end{tabular}} &
  \textbf{\begin{tabular}[c]{@{}c@{}}54.31\\ {[}48.98, 59.64{]}\end{tabular}} &
  \begin{tabular}[c]{@{}c@{}}60.02\\ {[}53.55, 69.77{]}\end{tabular} &
  \textbf{\begin{tabular}[c]{@{}c@{}}56.24\\ {[}48.84, 63.02{]}\end{tabular}} &
  \textbf{\begin{tabular}[c]{@{}c@{}}58.38\\ {[}51.78, 64.97{]}\end{tabular}} &
  \textbf{\begin{tabular}[c]{@{}c@{}}54.00\\ {[}47.70, 62.20{]}\end{tabular}} &
  \textbf{\begin{tabular}[c]{@{}c@{}}50.37\\ {[}43.14, 57.31{]}\end{tabular}} &
  \textbf{\begin{tabular}[c]{@{}c@{}}50.25\\ {[}43.15, 57.36{]}\end{tabular}} \\ \hline
\end{tabular}%
}
\end{table}




\clearpage
\section{GradCAM results} \label{ch7_app:cam}

The silver-standard PVS mask for the T2w image used in this analysis is shown in Figure \ref{ch7_fig:cam0_main} of the main body of the paper.

\subsection{First layer features}

\begin{figure}[htbp]
    \centering
    \includegraphics[width=1.0\linewidth]{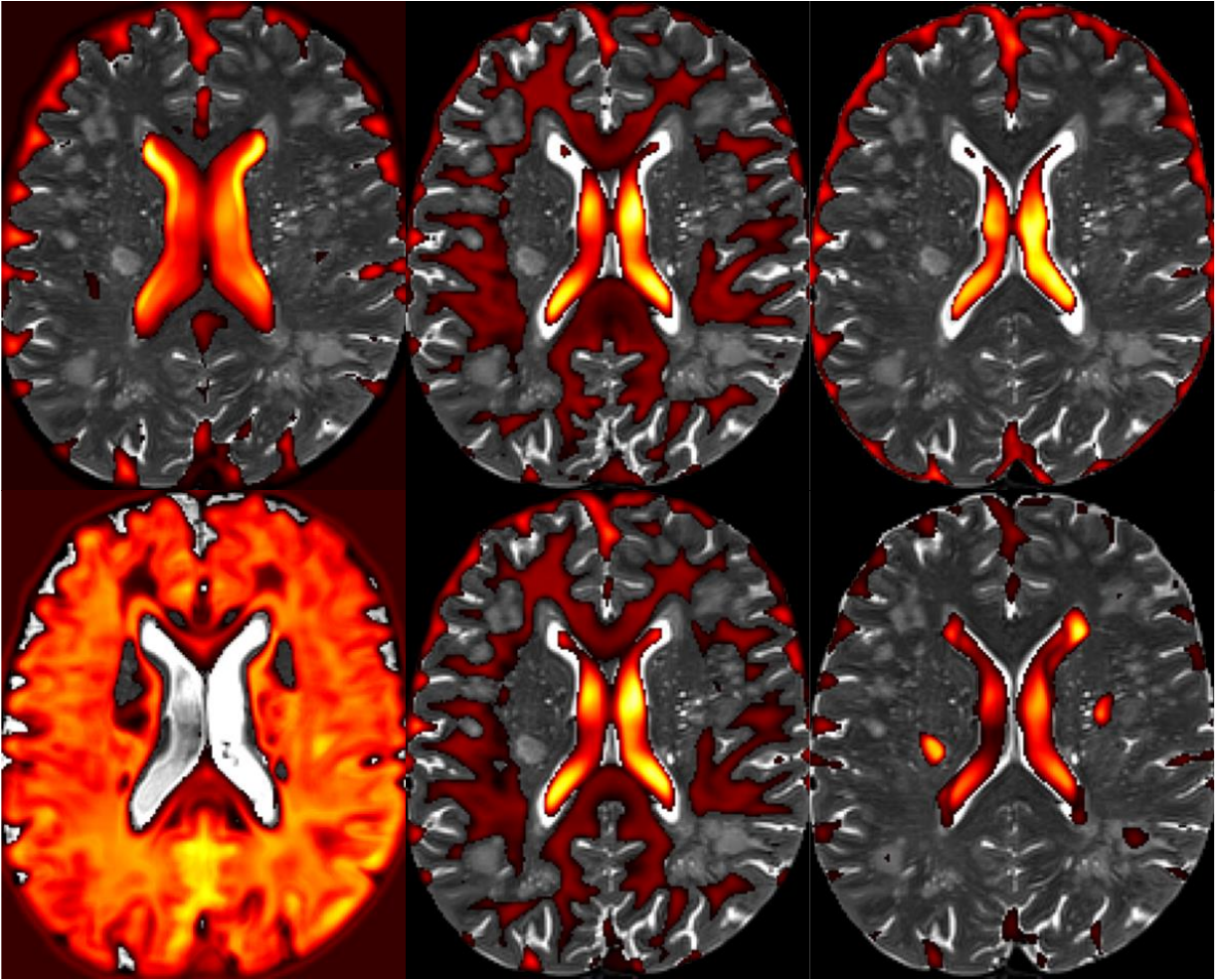}
    \caption{\textbf{Baseline CNN's GradCAM --- first layer}. Gradient-weighted class activation maps (GradCAM) for the first layer. The target class is perivascular space (PVS) score 3 for the basal ganglia (BG) (top) and centrum semiovale (CSO) (bottom). Results shown for three different training runs (left-right). There are no noticeable PVS-related activations. CNN: convolutional neural network.}
    \label{ch7_fig:cam0_app_baseline}
\end{figure}

\begin{figure}[htbp]
    \centering
    \includegraphics[width=1.0\linewidth]{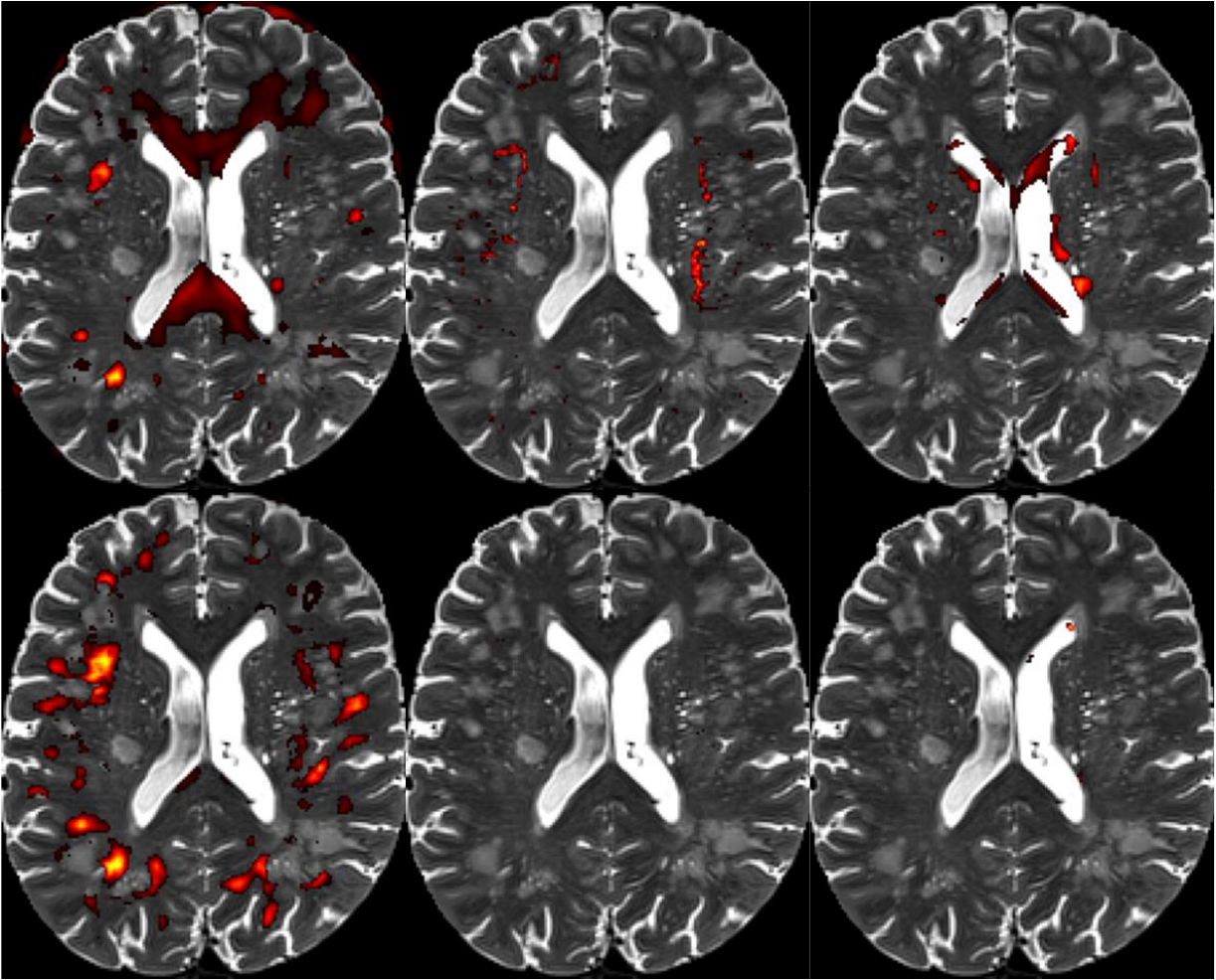}
    \caption{\textbf{Conditional CNN's GradCAM --- first layer}. Gradient-weighted class activation maps (GradCAM) for the first layer. The target class is perivascular space (PVS) score 3 for the basal ganglia (BG) (top) and centrum semiovale (CSO) (bottom). Results shown for three different training runs (left-right). There are no noticeable PVS-related activations. CNN: convolutional neural network.}
    \label{ch7_fig:cam0_app_conditional}
\end{figure}

\begin{figure}[htbp]
    \centering
    \includegraphics[width=1.0\linewidth]{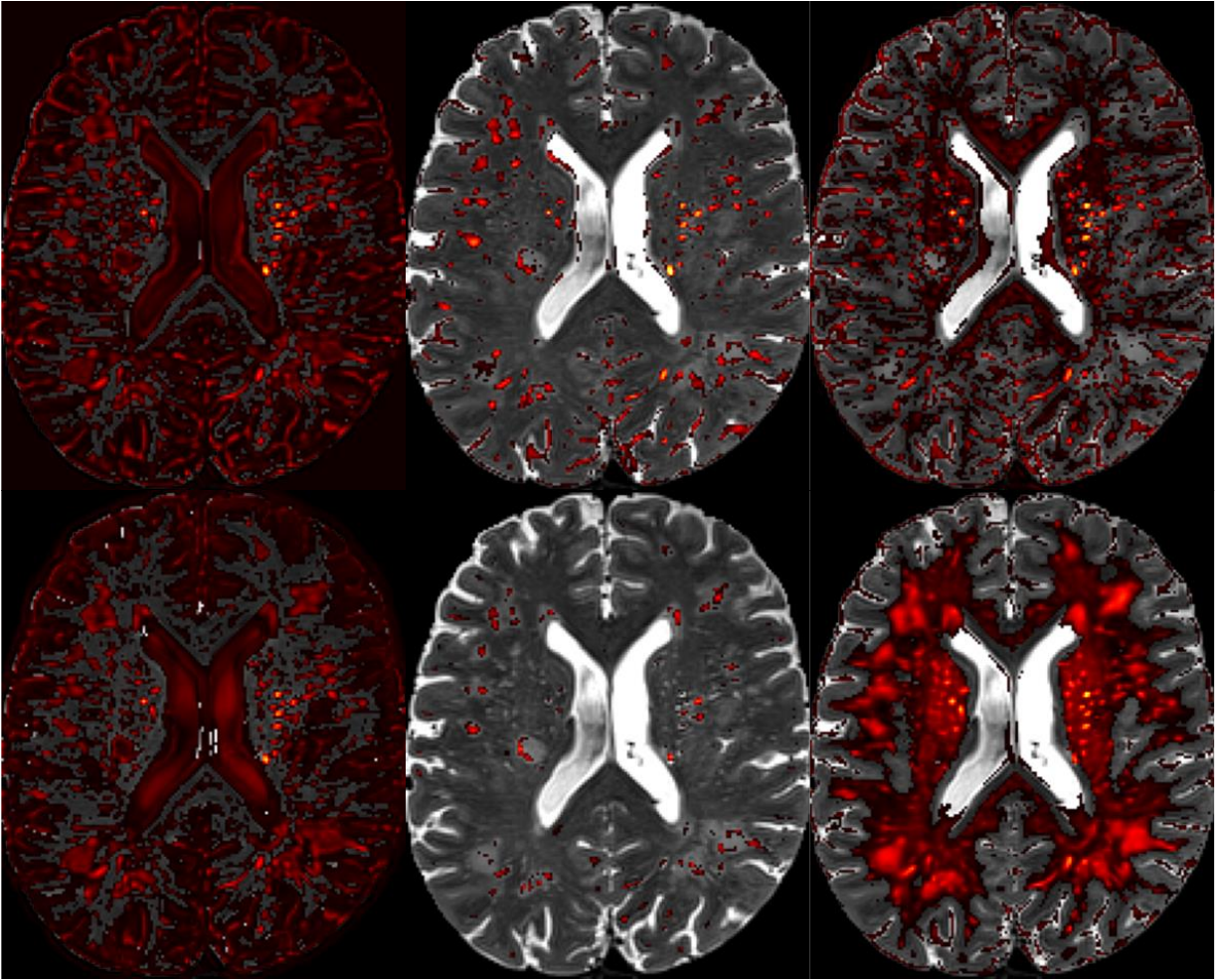}
    \caption{\textbf{Multi-task CNN's GradCAM --- first layer}. Gradient-weighted class activation maps (GradCAM) for the first layer. The target class is perivascular space (PVS) score 3 for the basal ganglia (BG) (top) and centrum semiovale (CSO) (bottom). Results shown for three different training runs (left-right). The highest activations in all images are overlapping with PVS. Both BG and CSO PVS are activated for both BG and CSO score 3 targets. This suggests that PVS in both regions might be considered in the prediction of the scores for both regions. CNN: convolutional neural network.}
    \label{ch7_fig:cam0_app_multitask}
\end{figure}

\clearpage
\subsection{Final layer features}

\begin{figure}[htbp]
    \centering
    \includegraphics[width=1.0\linewidth]{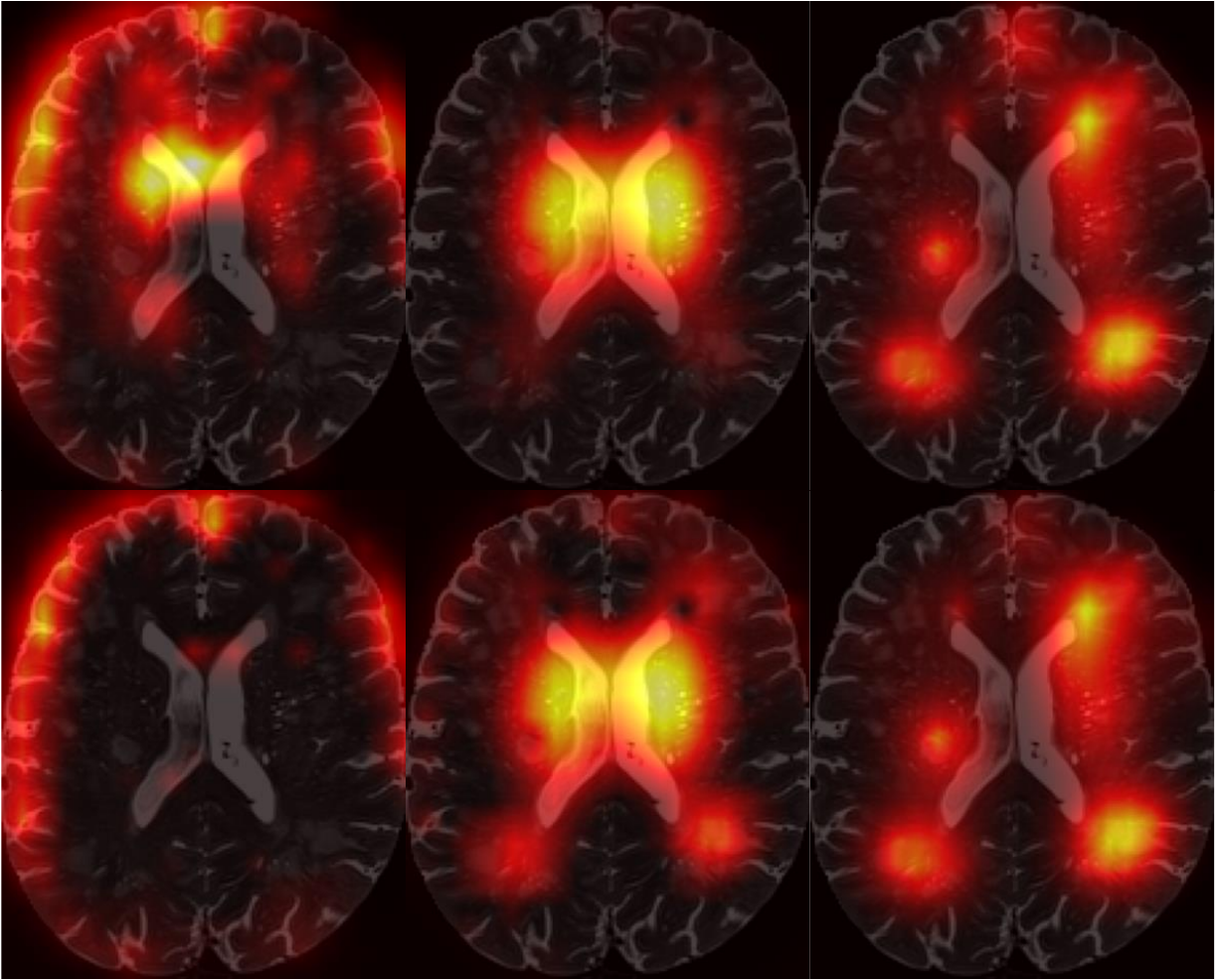}
    \caption{\textbf{Baseline CNN's GradCAM --- final layer}. Gradient-weighted class activation maps (GradCAM) for the final layer. The target class is perivascular space (PVS) score 3 for the basal ganglia (BG) (top) and centrum semiovale (CSO) (bottom). Results shown for three different training runs (left-right). Middle column shows strong activation in BG for both targets. Overall these maps are not very interpretable. CNN: convolutional neural network.}
    \label{ch7_fig:cam3_app_baseline}
\end{figure}

\begin{figure}[htbp]
    \centering
    \includegraphics[width=1.0\linewidth]{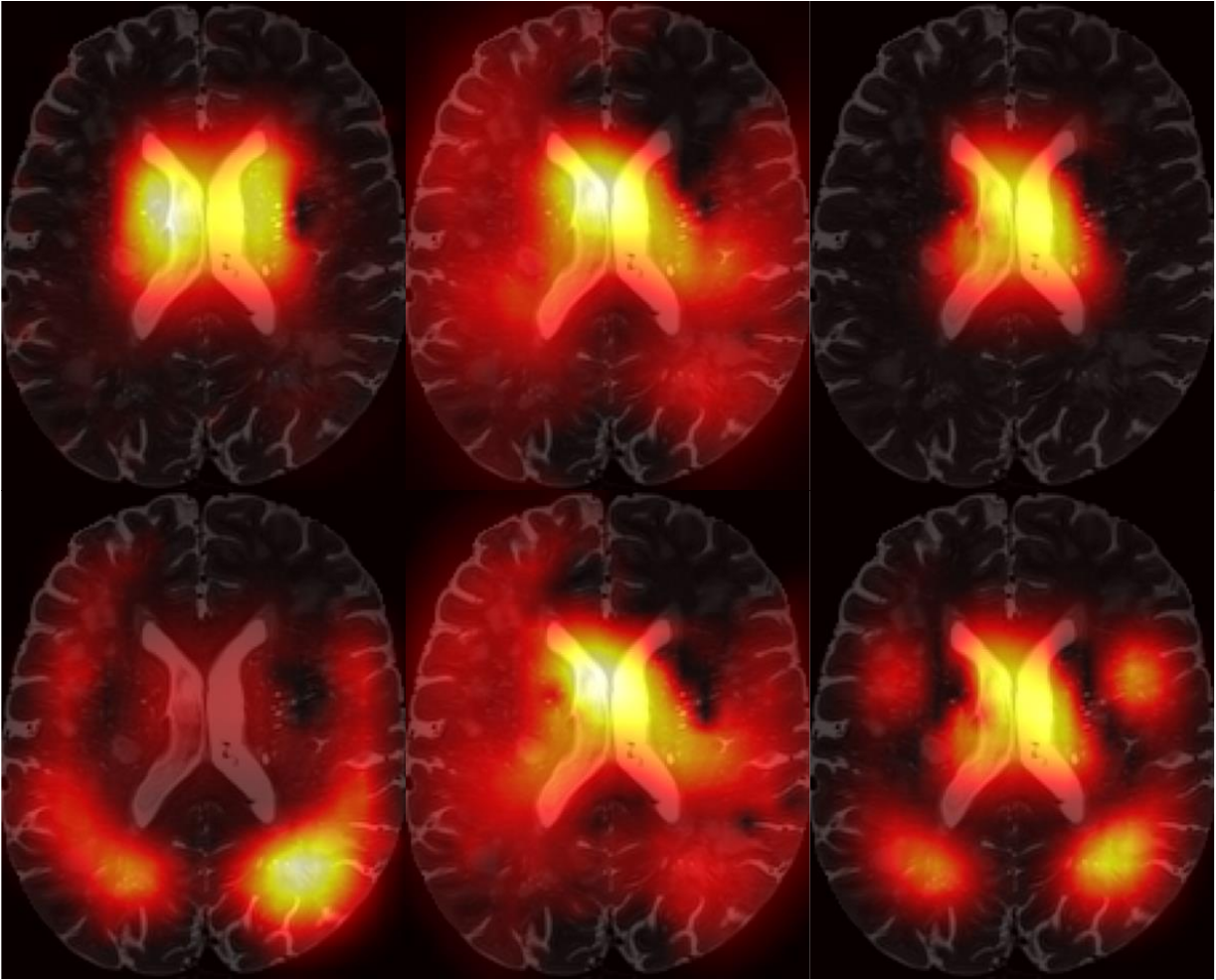}
    \caption{\textbf{Conditional CNN's GradCAM --- final layer}. Gradient-weighted class activation maps (GradCAM) for the final layer. The target class is perivascular space (PVS) score 3 for the basal ganglia (BG) (top) and centrum semiovale (CSO) (bottom). Results shown for three different training runs (left-right). All columns show strong activation in BG for the BG target. Columns 1 and 3 show more activation outside the BG for the CSO target than for the BG target. CNN: convolutional neural network.}
    \label{ch7_fig:cam3_app_conditional}
\end{figure}

\begin{figure}[htbp]
    \centering
    \includegraphics[width=1.0\linewidth]{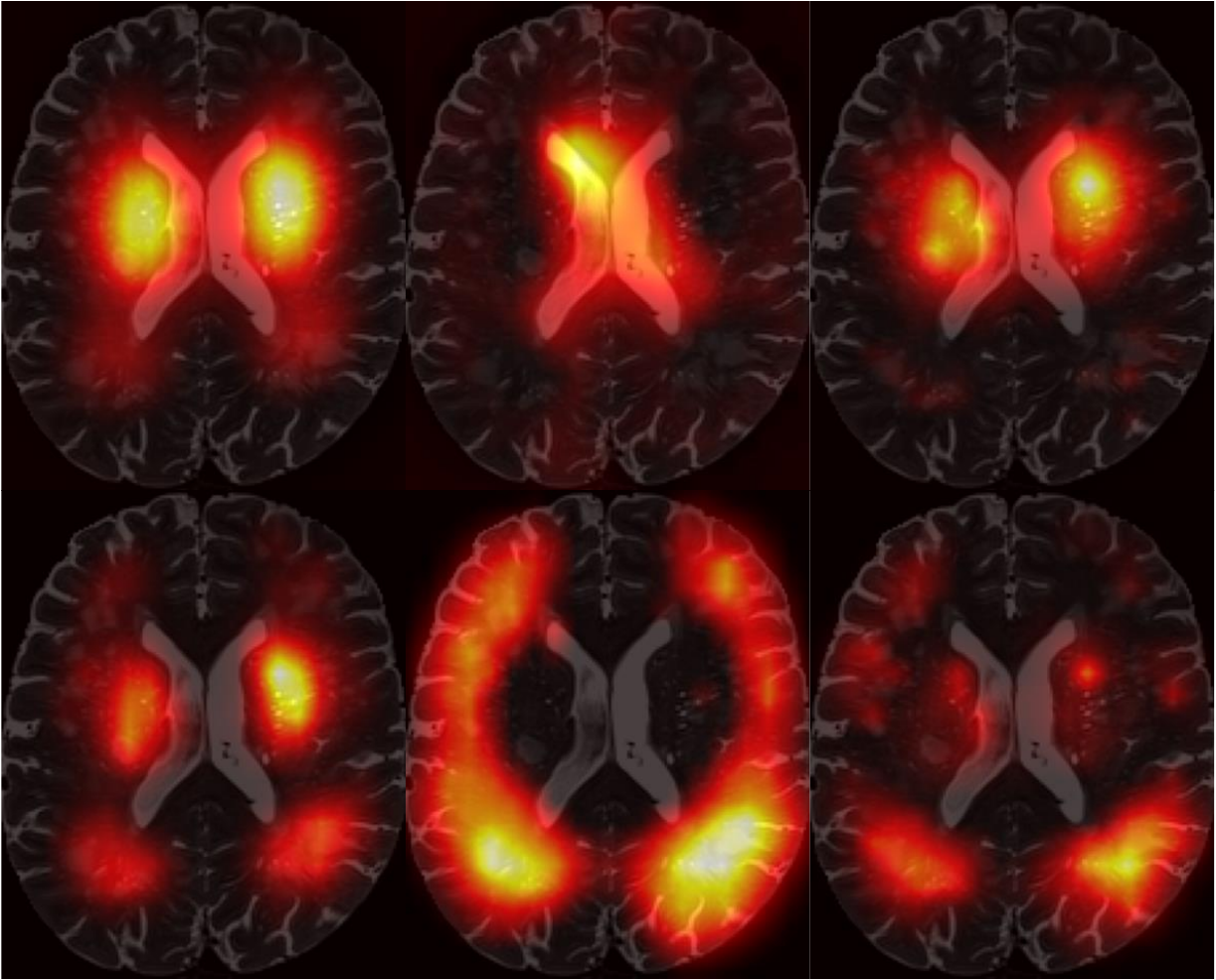}
    \caption{\textbf{Multi-task CNN's GradCAM --- final layer}. Gradient-weighted class activation maps (GradCAM) for the final layer. The target class is perivascular space (PVS) score 3 for the basal ganglia (BG) (top) and centrum semiovale (CSO) (bottom). Results shown for three different training runs (left-right). All columns show strong activation in BG for the BG target. All columns show more activation outside the BG for the CSO target than for the BG target. CNN: convolutional neural network.}
    \label{ch7_fig:cam3_app_multitask}
\end{figure}

\clearpage
\section{Pairwise statistical testing results} \label{ch7_app:stats}

Statistical testing was conducted via paired difference bootstrapping with 10,000 iterations and Bonferroni correction ($n=6$) was applied to control for multiple comparisons between models.

\subsection{Main method comparison}

\includepdf[pages=-, scale=0.9, offset=0 60, pagecommand={\thispagestyle{plain}}]{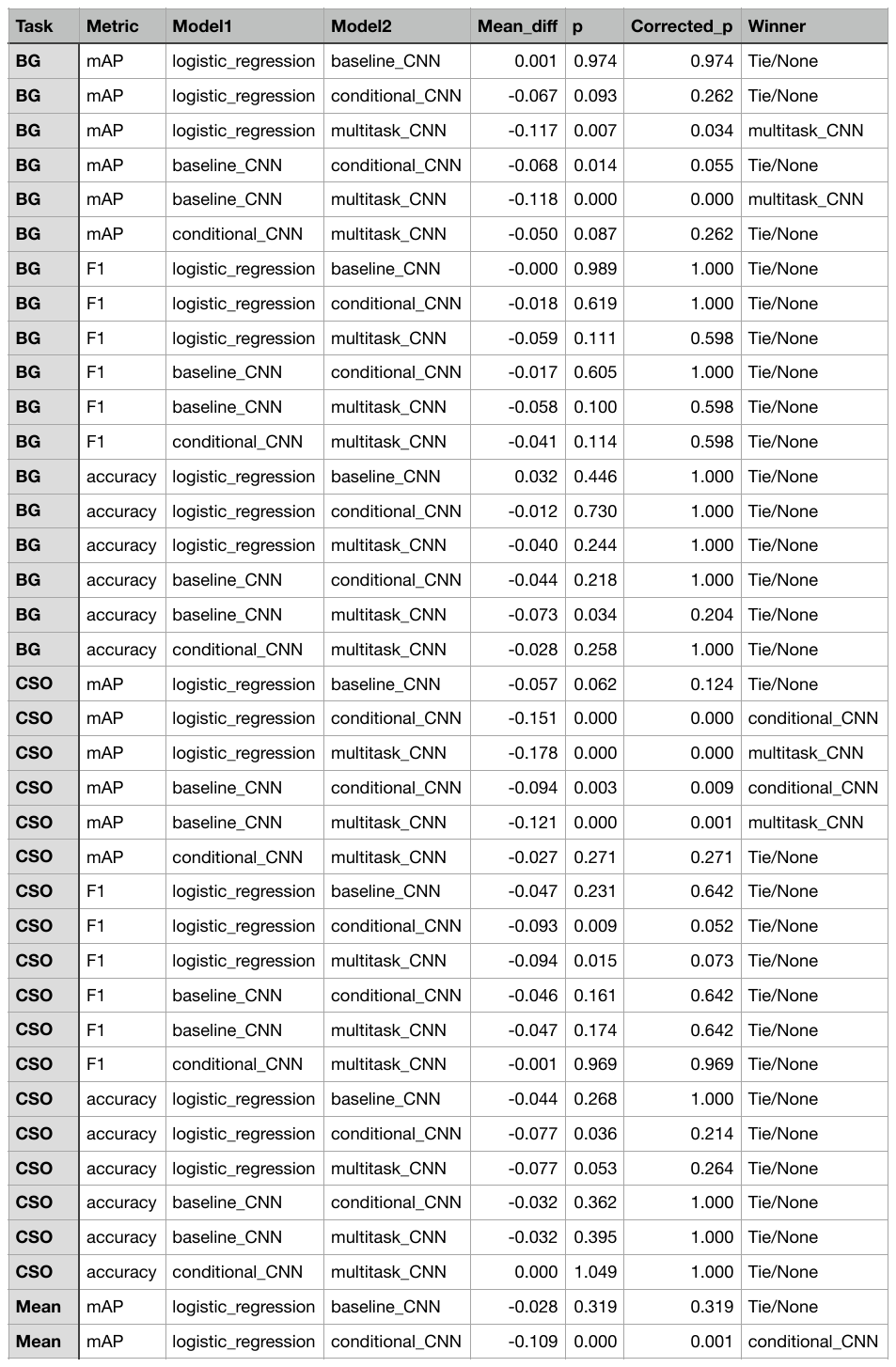}

\subsection{Original vs pseudo-label ablations}

\subsubsection{Logistic regression}

\includepdf[pages=-, scale=0.9, offset=0 60, pagecommand={\thispagestyle{plain}}]{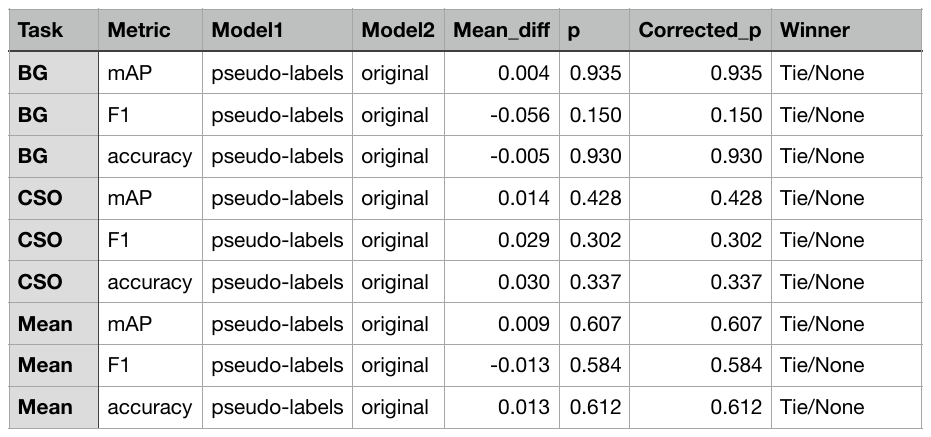}

\subsubsection{Conditional CNN}

\includepdf[pages=-, scale=0.9, offset=0 60, pagecommand={\thispagestyle{plain}}]{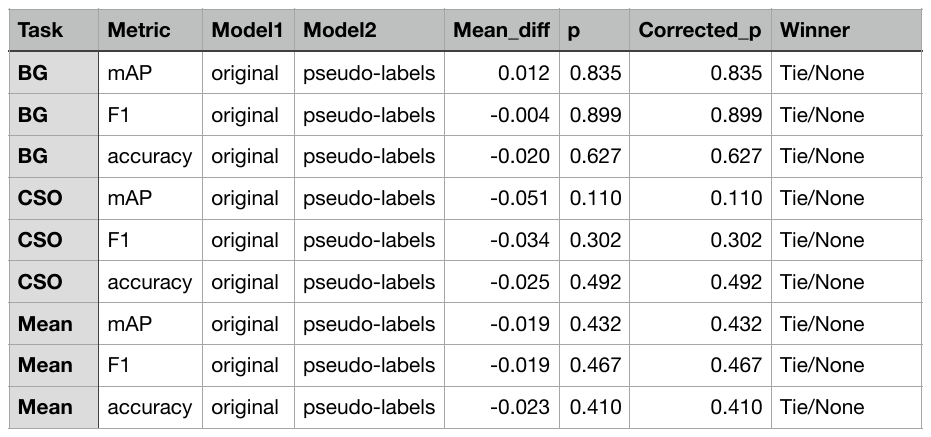}

\bibliographystyle{elsarticle-num} 
\bibliography{references}

@article{tillin2012southall,
  title={Southall And Brent REvisited: Cohort profile of SABRE, a UK population-based comparison of cardiovascular disease and diabetes in people of European, Indian Asian and African Caribbean origins},
  author={Tillin, Therese and Forouhi, Nita G and McKeigue, Paul M and Chaturvedi, Nish},
  journal={International journal of epidemiology},
  volume={41},
  number={1},
  pages={33--42},
  year={2012},
  publisher={Oxford University Press}
}

@article{jones2020cohort,
  title={Cohort Profile Update: Southall and Brent Revisited (SABRE) study: a UK population-based comparison of cardiovascular disease and diabetes in people of European, South Asian and African Caribbean heritage},
  author={Jones, Siana and Tillin, Therese and Park, Chloe and Williams, Suzanne and Rapala, Alicja and Al Saikhan, Lamia and Eastwood, Sophie V and Richards, Marcus and Hughes, Alun D and Chaturvedi, Nishi},
  journal={International journal of epidemiology},
  volume={49},
  number={5},
  pages={1441--1442e},
  year={2020},
  publisher={Oxford University Press}
}

@article{isensee2021nnu,
  title={nnU-Net: a self-configuring method for deep learning-based biomedical image segmentation},
  author={Isensee, Fabian and Jaeger, Paul F and Kohl, Simon AA and Petersen, Jens and Maier-Hein, Klaus H},
  journal={Nature methods},
  volume={18},
  number={2},
  pages={203--211},
  year={2021},
  publisher={Nature Publishing Group}
}

@article{potter2015cerebral,
    author = {Potter, Gillian M. and Chappell, Francesca M. and Morris, Zoe and Wardlaw, Joanna M.},
    title = {Cerebral Perivascular Spaces Visible on Magnetic Resonance Imaging: Development of a Qualitative Rating Scale and its Observer Reliability},
    journal = {Cerebrovascular Diseases},
    volume = {39},
    number = {3-4},
    pages = {224-231},
    year = {2015},
    month = {03},
    issn = {1015-9770},
    doi = {10.1159/000375153},
    url = {https://doi.org/10.1159/000375153},
    eprint = {https://karger.com/ced/article-pdf/39/3-4/224/2352977/000375153.pdf},
}

@inproceedings{ronneberger2015u,
  title={U-net: Convolutional networks for biomedical image segmentation},
  author={Ronneberger, Olaf and Fischer, Philipp and Brox, Thomas},
  booktitle={International Conference on Medical image computing and computer-assisted intervention},
  pages={234--241},
  year={2015},
  organization={Springer}
}

@inproceedings{he2016deep,
  title={Deep residual learning for image recognition},
  author={He, Kaiming and Zhang, Xiangyu and Ren, Shaoqing and Sun, Jian},
  booktitle={Proceedings of the IEEE conference on computer vision and pattern recognition},
  pages={770--778},
  year={2016}
}

@article{dice1945measures,
  title={Measures of the amount of ecologic association between species},
  author={Dice, Lee R},
  journal={Ecology},
  volume={26},
  number={3},
  pages={297--302},
  year={1945},
  publisher={JSTOR}
}

@article{wardlaw2009lacunar,
  title={Lacunar stroke is associated with diffuse blood--brain barrier dysfunction},
  author={Wardlaw, Joanna M and Doubal, Fergus and Armitage, Paul and Chappell, Francesca and Carpenter, Trevor and Mu{\~n}oz Maniega, Susana and Farrall, Andrew and Sudlow, Cathie and Dennis, Martin and Dhillon, Baljean},
  journal={Annals of Neurology: Official Journal of the American Neurological Association and the Child Neurology Society},
  volume={65},
  number={2},
  pages={194--202},
  year={2009},
  publisher={Wiley Online Library}
}

@article{waymont2024systematic,
  title={Systematic review and meta-analysis of automated methods for quantifying enlarged perivascular spaces in the brain},
  author={Waymont, Jennifer MJ and Hernandez, Maria del C Valdes and Bernal, Jose and Coello, Roberto Duarte and Brown, Rosalind and Chappell, Francesca M and Ballerini, Lucia and Wardlaw, Joanna M},
  journal={Neuroimage},
  volume={297},
  pages={120685},
  year={2024},
  publisher={Elsevier}
}

@article{wardlaw2017blood,
  title={Blood-brain barrier failure as a core mechanism in cerebral small vessel disease and dementia: evidence from a cohort study},
  author={Wardlaw, Joanna M and Makin, Stephen J and Hern{\'a}ndez, Maria C Vald{\'e}s and Armitage, Paul A and Heye, Anna K and Chappell, Francesca M and Munoz-Maniega, Susana and Sakka, Eleni and Shuler, Kirsten and Dennis, Martin S and others},
  journal={Alzheimer's \& Dementia},
  volume={13},
  number={6},
  pages={634--643},
  year={2017},
  publisher={Elsevier}
}

@article{clancy2021rationale,
  title={Rationale and design of a longitudinal study of cerebral small vessel diseases, clinical and imaging outcomes in patients presenting with mild ischaemic stroke: Mild Stroke Study 3},
  author={Clancy, Una and Garcia, Daniela Jaime and Stringer, Michael S and Thrippleton, Michael J and Vald{\'e}s-Hern{\'a}ndez, Maria C and Wiseman, Stewart and Hamilton, Olivia KL and Chappell, Francesca M and Brown, Rosalind and Blair, Gordon W and others},
  journal={European stroke journal},
  volume={6},
  number={1},
  pages={81--88},
  year={2021},
  publisher={SAGE Publications Sage UK: London, England}
}

@article{wardlaw2011brain,
  title={Brain aging, cognition in youth and old age and vascular disease in the Lothian Birth Cohort 1936: rationale, design and methodology of the imaging protocol},
  author={Wardlaw, Joanna M and Bastin, Mark E and Vald{\'e}s Hern{\'a}ndez, Maria C and Maniega, Susana Mu{\~n}oz and Royle, Natalie A and Morris, Zoe and Clayden, Jonathan D and Sandeman, Elaine M and Eadie, Elizabeth and Murray, Catherine and others},
  journal={International Journal of Stroke},
  volume={6},
  number={6},
  pages={547--559},
  year={2011},
  publisher={SAGE Publications Sage UK: London, England}
}

@article{sudre_where_2024,
  title={Where is VALDO? VAscular lesions detection and segmentation challenge at MICCAI 2021},
  author={Sudre, Carole H and Van Wijnen, Kimberlin and Dubost, Florian and Adams, Hieab and Atkinson, David and Barkhof, Frederik and Birhanu, Mahlet A and Bron, Esther E and Camarasa, Robin and Chaturvedi, Nish and others},
  journal={Med. Image Anal.},
  pages={103029},
  year={2023},
  publisher={Elsevier}
}

@article{hernandez2024influence,
  title={Influence of threshold selection and image sequence in in-vivo segmentation of enlarged perivascular spaces},
  author={Hernandez, Maria del C Valdes and Coello, Roberto Duarte and Xu, William and Bernal, Jos{\'e} and Cheng, Yajun and Ballerini, Lucia and Wiseman, Stewart J and Chappell, Francesca M and Clancy, Una and Garcia, Daniela Jaime and others},
  journal={Journal of neuroscience methods},
  volume={403},
  pages={110037},
  year={2024},
  publisher={Elsevier}
}

@article{valdes2023step,
  title={Step-by-step pipeline for segmenting enlarged perivascular spaces from 3D T2-weighted MRI, 2018-2023 [software]},
  author={Vald{\'e}s Hern{\'a}ndez, MDC and Ballerini, L and Glatz, A and Aribisala, BS and Bastin, ME and Dickie, DA and Duarte Coello, R and Munoz Maniega, S and Wardlaw, JM},
  journal={University of Edinburgh. College of Medicine and Veterinary Medicine. Centre for Clinical Brain Sciences. DOI: https://doi. org/10.7488/ds/7486},
  year={2023}
}

@article{hoopes2022synthstrip,
  title={SynthStrip: skull-stripping for any brain image},
  author={Hoopes, Andrew and Mora, Jocelyn S and Dalca, Adrian V and Fischl, Bruce and Hoffmann, Malte},
  journal={NeuroImage},
  volume={260},
  pages={119474},
  year={2022},
  publisher={Elsevier}
}

@article{billot2023synthseg,
  title={SynthSeg: Segmentation of brain MRI scans of any contrast and resolution without retraining},
  author={Billot, Benjamin and Greve, Douglas N and Puonti, Oula and Thielscher, Axel and Van Leemput, Koen and Fischl, Bruce and Dalca, Adrian V and Iglesias, Juan Eugenio and others},
  journal={Medical image analysis},
  volume={86},
  pages={102789},
  year={2023},
  publisher={Elsevier}
}

@article{modat2014global,
  title={Global image registration using a symmetric block-matching approach},
  author={Modat, Marc and Cash, David M and Daga, Pankaj and Winston, Gavin P and Duncan, John S and Ourselin, S{\'e}bastien},
  journal={Journal of medical imaging},
  volume={1},
  number={2},
  pages={024003--024003},
  year={2014},
  publisher={Society of Photo-Optical Instrumentation Engineers}
}

@article{modat2010fast,
  title={Fast free-form deformation using graphics processing units},
  author={Modat, Marc and Ridgway, Gerard R and Taylor, Zeike A and Lehmann, Manja and Barnes, Josephine and Hawkes, David J and Fox, Nick C and Ourselin, S{\'e}bastien},
  journal={Computer methods and programs in biomedicine},
  volume={98},
  number={3},
  pages={278--284},
  year={2010},
  publisher={Elsevier}
}

@article{perez2021torchio,
  title={TorchIO: a Python library for efficient loading, preprocessing, augmentation and patch-based sampling of medical images in deep learning},
  author={P{\'e}rez-Garc{\'\i}a, Fernando and Sparks, Rachel and Ourselin, S{\'e}bastien},
  journal={Computer methods and programs in biomedicine},
  volume={208},
  pages={106236},
  year={2021},
  publisher={Elsevier}
}

@article{cardoso2022monai,
  title={Monai: An open-source framework for deep learning in healthcare},
  author={Cardoso, M Jorge and Li, Wenqi and Brown, Richard and Ma, Nic and Kerfoot, Eric and Wang, Yiheng and Murrey, Benjamin and Myronenko, Andriy and Zhao, Can and Yang, Dong and others},
  journal={arXiv preprint arXiv:2211.02701},
  year={2022}
}

@article{wardlaw2020perivascular,
  title={Perivascular spaces in the brain: anatomy, physiology and pathology},
  author={Wardlaw, Joanna M and Benveniste, Helene and Nedergaard, Maiken and Zlokovic, Berislav V and Mestre, Humberto and Lee, Hedok and Doubal, Fergus N and Brown, Rosalind and Ramirez, Joel and MacIntosh, Bradley J and others},
  journal={Nature Reviews Neurology},
  volume={16},
  number={3},
  pages={137--153},
  year={2020},
  publisher={Nature Publishing Group UK London}
}

@article{selvaraju2020grad,
  title={Grad-CAM: visual explanations from deep networks via gradient-based localization},
  author={Selvaraju, Ramprasaath R and Cogswell, Michael and Das, Abhishek and Vedantam, Ramakrishna and Parikh, Devi and Batra, Dhruv},
  journal={International journal of computer vision},
  volume={128},
  number={2},
  pages={336--359},
  year={2020},
  publisher={Springer}
}

@article{gonzalez2017reliability,
  title={Reliability of an automatic classifier for brain enlarged perivascular spaces burden and comparison with human performance},
  author={Gonzalez-Castro, Victor and Vald{\'e}s Hern{\'a}ndez, Mar{\'\i}a del C and Chappell, Francesca M and Armitage, Paul A and Makin, Stephen and Wardlaw, Joanna M},
  journal={Clinical Science},
  volume={131},
  number={13},
  pages={1465--1481},
  year={2017},
  publisher={Portland Press Ltd.}
}

@article{ballerini2018perivascular,
  title={Perivascular spaces segmentation in brain MRI using optimal 3D filtering},
  author={Ballerini, Lucia and Lovreglio, Ruggiero and Vald{\'e}s Hern{\'a}ndez, Maria del C and Ramirez, Joel and MacIntosh, Bradley J and Black, Sandra E and Wardlaw, Joanna M},
  journal={Scientific reports},
  volume={8},
  number={1},
  pages={2132},
  year={2018},
  publisher={Nature Publishing Group UK London}
}

@article{adams2013rating,
  title={Rating method for dilated Virchow--Robin spaces on magnetic resonance imaging},
  author={Adams, Hieab HH and Cavalieri, Margherita and Verhaaren, Benjamin FJ and Bos, Daniel and van der Lugt, Aad and Enzinger, Christian and Vernooij, Meike W and Schmidt, Reinhold and Ikram, M Arfan},
  journal={Stroke},
  volume={44},
  number={6},
  pages={1732--1735},
  year={2013},
  publisher={Lippincott Williams \& Wilkins Hagerstown, MD}
}

@article{dubost2019enlarged,
  title={Enlarged perivascular spaces in brain MRI: automated quantification in four regions},
  author={Dubost, Florian and Yilmaz, Pinar and Adams, Hieab and Bortsova, Gerda and Ikram, M Arfan and Niessen, Wiro and Vernooij, Meike and de Bruijne, Marleen},
  journal={Neuroimage},
  volume={185},
  pages={534--544},
  year={2019},
  publisher={Elsevier}
}

@article{yang2021direct,
  title={Direct rating estimation of enlarged perivascular spaces (Epvs) in brain MRI using deep neural network},
  author={Yang, Ehwa and Gonuguntla, Venkateswarlu and Moon, Won-Jin and Moon, Yeonsil and Kim, Hee-Jin and Park, Mina and Kim, Jae-Hun},
  journal={Applied Sciences},
  volume={11},
  number={20},
  pages={9398},
  year={2021},
  publisher={MDPI}
}

@article{williamson2022automated,
  title={Automated grading of enlarged perivascular spaces in clinical imaging data of an acute stroke cohort using an interpretable, 3D deep learning framework},
  author={Williamson, Brady J and Khandwala, Vivek and Wang, David and Maloney, Thomas and Sucharew, Heidi and Horn, Paul and Haverbusch, Mary and Alwell, Kathleen and Gangatirkar, Shantala and Mahammedi, Abdelkader and others},
  journal={Scientific Reports},
  volume={12},
  number={1},
  pages={788},
  year={2022},
  publisher={Nature Publishing Group UK London}
}

@article{taylor2018cohort,
  title={Cohort profile update: the Lothian Birth Cohorts of 1921 and 1936},
  author={Taylor, Adele M and Pattie, Alison and Deary, Ian J},
  journal={International journal of epidemiology},
  volume={47},
  number={4},
  pages={1042--1042r},
  year={2018},
  publisher={Oxford University Press}
}

\end{document}